\documentclass{article}

\usepackage{microtype}
\usepackage{graphicx}
\usepackage{subcaption}
\usepackage{booktabs} 
\usepackage[export]{adjustbox}

\usepackage{hyperref}

\usepackage{graphicx}
\graphicspath{ {images/} }
\usepackage{amsmath, amssymb}
\usepackage[shortlabels]{enumitem}
\renewcommand{\eqref}[1]{(\ref{#1})}
\usepackage{pdfpages}
\usepackage{wrapfig,lipsum,booktabs}
\usepackage{balance}

\usepackage[accepted]{icml2021}

\icmltitlerunning{Exploration in Approximate Hyper-State Space}

\begin{document}

\twocolumn[
\icmltitle{Exploration in Approximate Hyper-State Space \\ for Meta Reinforcement Learning}



\icmlsetsymbol{equal}{*}

\begin{icmlauthorlist}
\icmlauthor{Luisa Zintgraf}{to}
\icmlauthor{Leo Feng}{goo}
\icmlauthor{Cong Lu}{to}
\icmlauthor{Maximilian Igl}{to}
\icmlauthor{Kristian Hartikainen}{to}
\\
\icmlauthor{Katja Hofmann}{ed}
\icmlauthor{Shimon Whiteson}{to}
\end{icmlauthorlist}

\icmlaffiliation{to}{University of Oxford, UK.}
\icmlaffiliation{goo}{Mila, Université de Montréal, Canada.}
\icmlaffiliation{ed}{Microsoft Research, Cambridge, UK}

\icmlcorrespondingauthor{Luisa Zintgraf}{luisa.zintgraf@cs.ox.ac.uk}

\icmlkeywords{meta learning, sparse reward, bayes optimal, BAMDP}

\vskip 0.3in
]



\printAffiliationsAndNotice{}  

\begin{abstract}
	To rapidly learn a new task, it is often essential for agents to explore efficiently -- especially when performance matters from the first timestep. 
	One way to learn such behaviour is via meta-learning. 
	Many existing methods however rely on dense rewards for meta-training, and can fail catastrophically if the rewards are sparse. 
	Without a suitable reward signal, the need for exploration \emph{during meta-training} is exacerbated. 
	To address this, 
	we propose HyperX,
	which uses novel reward bonuses for meta-training to explore in \emph{approximate hyper-state space} (where hyper-states represent the environment state and the agent's task belief). 
	We show empirically that HyperX meta-learns better task-exploration and adapts more successfully to new tasks than existing methods.
\end{abstract}


\section{Introduction}

\begin{figure}
	\centering
	\includegraphics[width=0.85\columnwidth]{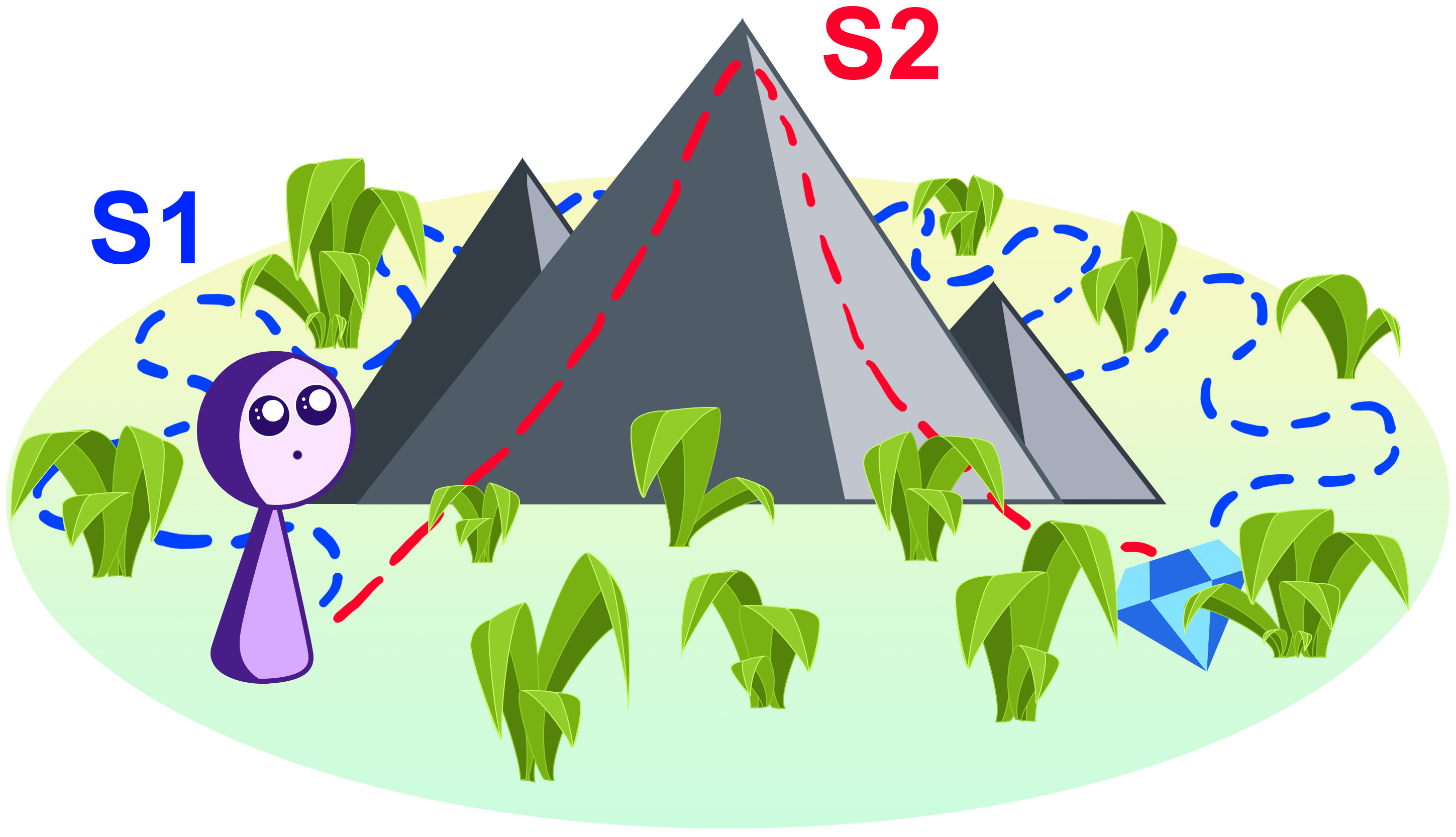}
	\caption{
		\textbf{Illustration of the Meta-Exploration Problem.} 
		In this environment, the agent must find the hidden treasure but cannot see above the grass. 
		Two possible \emph{task-exploration} strategies are to (s1) search the grass for the treasure, or (s2) climb up the mountain, see the treasure, and go there. 
		The agent is meta-trained across different treasure locations, and should try both s1 and s2 to find out which one leads to higher online return in expectation across tasks. 
		We call this \emph{meta-exploration} across task-exploration strategies.
		Due to a lack of sufficient meta-exploration, existing methods often fail to find the (superior) task-exploration strategy s2 (Sec \ref{sec:experiments:treasure}). 
	}
	\label{fig:comic_treasure}
\end{figure}

In meta-learning, the experience of a machine learning model over multiple learning episodes (which can range from single model updates to entire lifetimes), 
is used to improve future \emph{learning} performance \citep{hospedales2020meta}, e.g., in terms of data or computational efficiency.
This ``learning to learn'' requires a feedback signal on the meta-level, which quantifies how well the model performs after a learning episode.
In meta reinforcement learning (meta-RL), this is often measured by episodic (online or final) return of the agent, or learning improvement.
If this feedback signal is not present because the agent has not made sufficient learning progress, meta-learning can fail to solve the task.
In this case, the agent faces an exploration challenge at the meta-level: how to find a good meta-learning signal in the first place.

A weakness of existing meta-RL methods is that they often rely on dense rewards 
or sufficiently small state spaces,
such that even with naive exploration during meta-training the agent receives rewards that guide it towards good behaviour.
However, we observe empirically (Sec \ref{sec:experiments}) that if the rewards are sparse or if exploratory behaviour is penalised in the short term, existing methods can fail.
This is problematic, since defining dense rewards can be tedious and error-prone, and many real-world applications have sparse rewards (e.g., a fail/success criterion).
Hence, to make meta-learning practical for such settings, we need methods that can meta-learn even when rewards are sparse. 

In this paper we consider this challenge in the context of meta-learning \emph{fast online adaptation} to tasks from a given distribution.
In this setting, the agent aims to maximise expected online return when adapting to an unknown task, which requires it to carefully trade off exploration and exploitation.
To make this distinction clear, we call this \emph{task-exploration} in contrast to \emph{meta-exploration}, illustrated with an example in Figure \ref{fig:comic_treasure}.
\textbf{Task-exploration} refers to the exploration behaviour we want to meta-learn: 
when in a new environment, the agent must explore to learn the task.
We want this agent to be Bayes-optimal, i.e., to maximise expected \emph{online} return that it incurs \emph{while} learning about the environment.
\textbf{Meta-exploration} refers to the challenge of \emph{exploring across tasks and adaptation behaviours} during meta-training.
The agent has to (a) explore across individual tasks since the same state can have different values across tasks, and (b) learn about the shared structure between tasks to extract information about how to adapt, 
i.e., the agent must try out different task-exploration strategies during meta-training to find the Bayes-optimal one.
Contrary to the Bayes-optimal task-exploration we want to meta-learn, we do \emph{not} care about the rewards incurred during meta-exploration, 
but rather about efficiently gathering the data needed for meta-learning.

The Bayes-optimal policy can in principle be found by solving a Bayesian formulation of the reinforcement learning (RL) problem that considers both the environment state and the agent's internal belief about the environment, together called \textit{hyper-states} \cite{duff2002optimal}.
This allows the agent to compute how to explore optimally under task uncertainty:
it takes information-seeking actions 
(which can be costly in the short term) 
\emph{if and only if}
they lead to higher expected long-term returns
(by yielding information that can be exploited later).
While this computation is intractable for all but the simplest environments, recent work makes
significant progress by \emph{meta-learning} approximately Bayes-optimal behaviour 
 for a given task distribution
\citep{duan2016rl, wang2016learning, ortega2019meta, humplik2019meta, zintgraf2020varibad, mikulik2020meta}. 

We propose HyperX (\textbf{Hyper}-State E\textbf{x}ploration), a novel method for meta-learning approximately Bayes-optimal exploration strategies when rewards are sparse.
Similar to VariBAD \citep{zintgraf2020varibad}, HyperX simultaneously meta-learns approximate task inference, and trains a policy that conditions on hyper-states, i.e., the environment state and the approximate task belief.
To ensure sufficient meta-exploration, HyperX combines two exploration bonuses during meta-training.
The first is a novelty bonus on approximate hyper-states using random network distillation \citep{osband2018randomized, burda2019exploration} that encourages the agent to try out different task-exploration strategies, so that it can better find an approximately Bayes-optimal one.
However, as this requires accurate task inference which we aim to meta-learn alongside the policy, the beliefs are inaccurate early in training and this bonus is not useful by itself. 
We therefore use a second exploration bonus to incentivise the agent to gather the data necessary to learn approximate belief inference. 
This bonus is computed using the discrepancy between the rewards and transitions that the belief model decoder predicts, and the ground-truth rewards and transitions the agent observes.
This exploration bonus encourages the agent to visit states where the belief inference is incorrect and more data should be collected.

We show empirically that in environments without dense and informative rewards, current state of the art methods either fail to learn, or learn sub-optimal adaptation behaviour. 
In contrast, we show that HyperX can successfully meta-learn approximately Bayes-optimal strategies on these tasks.


\section{Background}

Our goal is to meta-learn policies that maximise expected \textit{online} return, i.e., optimally trade off exploration and exploitation under task uncertainty. 
We formally define this problem setting below.

\subsection{Problem Setting}

\textbf{Task Distribution.} 
We consider a meta-learning setting where we have a distribution $p(M)$ over MDPs. 
An MDP $M_i {\sim} p(M)$ is defined by a tuple $M_i {=} (\mathcal{S}, \mathcal{A}, R_i,  T_i, \gamma, H)$.
$\mathcal{S}$ is a set of states, $\mathcal{A}$ a set of actions, $R(r_{t+1}|s_t, a_t, s_{t+1})$ a reward function, $T(s_{t+1}|s_t, a_t)$ a transition function including the initial state distribution $T_i(s_0)$, $\gamma$ a discount factor, and $H$ the horizon. 
Across tasks, the reward and transition functions can vary so we often express $p(M)$ as $p(R, T)$.

\textbf{Objective.} 
Our objective is to meta-learn a policy that, when deployed in an (unseen) test task drawn from $p(M)$, 
maximises the online return achieved \emph{during task-learning}:
$
	\max_{\pi}~\mathbb{E}_{p(M)}\left[ \mathcal{J}(\pi) \right]  
$
where
$\mathcal{J}(\pi)=\mathbb{E}_{T, R, \pi}[\sum_{t=0}^{H-1} \gamma^t r_t]$.
Since the agent does not initially know which MDP it is in, maximising this objective requires a good task-exploration strategy to cope with the initially unknown reward and transition functions, and to exploit task information to adapt in this environment.
The more an agent can make use of prior knowledge about $p$, the better it can perform this trade-off.

\textbf{Meta-Learning.} 
During meta-training we assume access to a task distribution $p(M)$, from which we sample batches of tasks $\mathbf{M}=\{M_i\}_{i=1}^N$ and interact with them to learn good task-exploration strategies. 
During this phase, we need good meta-exploration, to collect the data necessary for meta-learning.
At meta-test time, the agent is evaluated based on the expected return it gets \textit{while adapting} to new tasks from $p(M)$.
This requires good task-exploration strategies.

\subsection{Bayesian Reinforcement Learning.} \label{background:bamdp}

In principle, we can compute the optimal solution to the problem describe above by formulating the problem as a Bayes-Adaptive MDP (BAMDP, \citet{duff2002optimal}), 
which is a tuple $M^+ = \left( \mathcal{S}^+, \mathcal{A}, R^+, T^+, T^+_0, \gamma, H^+ \right)$. 
Here, $\mathcal{S}^+ = \mathcal{S} \times \mathcal{B} $ is the hyper-state space, consisting of the underlying MDP environment state space $\mathcal{S}$ and a belief space $\mathcal{B}$ whose elements are beliefs over the MDP.
This belief is typically expressed as a distribution over the reward and transition function $b_t(R, T)=p(R, T|\tau_{:_t})$, where $\tau_{:t}=(s_0, a_0, r_1, s_1, \dotsc, s_t)$ is the agent's experience up until the current time step $t$ in the current task.
The transition function is defined as 
$
T^+(s^+_{t+1} | s_t^+, a_t, r_t) = \nonumber \mathbb{E}_{b_t}\left[ T(s_{t+1} | s_t, a_t) \right] ~ \delta(b_{t+1}{=}p(R, T|\tau_{:t+1}))
\label{eq:bamdp_transition_function}
$
and the reward function as 
$
R^+(s_t^+, a_t, s^+_{t+1}) = \mathbb{E}_{b_{t+1}} \left[ R(s_t, a_t, s_{t+1}) \right] .
\label{eq:bamdp_reward_function}
$
$T^+_0(s^+)$ is the initial hyper-state distribution, and $H^+$ is the horizon in the BAMDP. 

A policy $\pi(s^+)$ acting in a BAMDP conditions its actions not only on the environment state $s$, but also on the belief $b$. 
This way, it can take task uncertainty into account when making decisions. 
The agent's objective in a BAMDP is to maximise the expected return in an initially unknown environment, while learning, within the horizon $H^+$:
\begin{equation} \label{eq:bamdp_objective}
\mathcal{J}^+(\pi) =
\mathbb{E}_{b_0,  T^+, \pi} \left[ \sum_{t=0}^{H^+-1} \gamma^t R^+(r_{t+1} | s^+_t, a_t, s^+_{t+1}) \right] .
\end{equation}
A policy $\pi(s_t, b_t)$ that maximises this objective is called Bayes-optimal, as it optimally trades off exploration and exploitation in order to maximise expected cumulative return.
For an in-depth introduction to BAMDPs, see \citet{duff2002optimal} or \citet{ghavamzadeh2015bayesian}.

The belief inference and planning in belief space is generally intractable, but we can meta-learn an approximate inference procedure \citep{ortega2019meta, mikulik2020meta}.
Existing methods meta-learn to maintain a belief either implicitly within the workings of recurrent networks (RL$^2$, \citet{duan2016rl, wang2016learning}), or explicitly by meta-learning a posterior using privileged  \citep{humplik2019meta} or unsupervised (VariBAD, \citet{zintgraf2020varibad}) information.
In this paper we use VariBAD, because it explicitly expresses the belief as a single latent vector, which we need in order to compute the exploration bonus.

\subsection{VariBAD}

VariBAD \citep{zintgraf2020varibad} jointly trains
a policy $\pi_\psi(s_t, b_t)$, and
a variational auto-encoder (VAE, \citet{kingma2013auto}) for approximate belief inference.
The VAE consists of 
an encoder $q_\theta(m|\tau_{:t})$ to compute an approximate belief $b_t$, 
and
reward and transition decoders $p(r_{i+1} | s_i, a_i, s_{i+1}, m_t)$ and $p(s_{i+1}  | s_i, a_i, m_t)$ with $m\sim b_t$ which are used only during meta-training.
The objective is
\begin{equation} \label{eq:varibad_objective}
\mathcal{L}(\phi,\theta,\psi) 
= \mathbb{E}_{p(M)} \left[\mathcal{J}(\psi) 
+ \sum_{t=0}^{H^+}
~ ELBO_t(\phi,\theta) 
~ \right] 
\end{equation}
where
\begin{align} \label{eq:elbos}
ELBO_t 
=~& \mathbb{E}_{p(M)} \left[  \mathbb{E}_{ q_\phi(m|\tau_{:t})} \left[ \log p_\theta(\tau_{:H^+}| m) \right] \right. \nonumber  \\
&- \left. KL(q_\phi(m|\tau_{:t}) || q_\phi(m|\tau_{:t-1})) \right]   ,
\end{align}
with prior $q_\phi(m) = \mathcal{N}(0, I)$.
The objective jointly maximises an RL loss $\mathcal{J}$ (with the agent conditioned on state $s_t$ and approximate belief $b_t$ represented by the mean and variance of the VAE's latent distribution) and an evidence lower bound (ELBO) on the environment model, that consists of a reconstruction term for the trajectory and a KL divergence to the previous posterior. 
Like \citet{zintgraf2020varibad}, we do not backpropagate the RL loss through the encoder (hence $\mathcal{J}$ does not depend on the encoder parameters $\phi$).


\section{Method: HyperX} \label{sec:method}

Meta-learning good task-adaptation behaviour requires the agent to, during meta-training, gather the data necessary to learn good task-exploration strategies.
If the environment rewards are sparse, they might however not provide enough signal for an agent to learn something if it follows naive exploration during meta-training. 
In that case, existing methods can fail and special attention needs to be paid to meta-exploration.
The agent needs to explore the state space sufficiently during meta-training, which is complicated by the fact that the same state can have different values across tasks.  
A good meta-exploration strategy also ensures that the agent tries out diverse task-exploration strategies which allow it to find an approximately Bayes-optimal one.

To address the meta-exploration problem, we propose HyperX (\textbf{Hyper}-State-E\textbf{x}ploration), a  method to meta-learn approximately Bayes-optimal behaviour even when rewards are not dense.
The two key ideas behind HyperX are:
\begin{enumerate}
	\item 
	We can incentivise the agent to try out different task-exploration strategies during meta-training by rewarding novel \emph{hyper}-states.
	By exploring the \emph{joint} space of beliefs and states (i.e., hyper-states), the agent simultaneously (a) explores the state space, while distinguishing between visitation counts in different tasks due to changing beliefs, and (b) tries out different task-exploration strategies because these lead to different beliefs (even in the same state).
	To achieve this, we add an exploration bonus $r^\text{hyper}(s^+)$ that rewards visiting novel hyper-states.
	\item 
	For the novelty bonus on hyper-states to be meaningful, the beliefs needs to be meaningful. 
	However since the inference procedure is meta-learned alongside the policy, they do not capture task information early in training.
	We therefore additionally incentivise the agent to explore states where beliefs are inaccurate, by using the VAE reconstruction error (of the current rewards and transitions given the current belief) as a reward bonus, $r^\text{error}(s_t, r_t)$. 
	Since the belief is conditioned on the history \emph{including} the most recent reward $r_t$ and state $s_t$, and the VAE is trained 
	to predict rewards and states given beliefs, this bonus tends to zero over training.
\end{enumerate}
In the following, we describe how to compute these bonuses.
 
\textbf{Hyper-State Exploration.}
To compute exploration bonuses on the hyper-states, we use random network distillation
(see Appendix \ref{sec:background:rpfs}) given its empirical successes in standard RL problems \citep{osband2017deep, osband2018randomized, burda2019exploration} and theoretical justifications for deep networks \citep{pearce2018uncertainty, ciosek2020conservative}.
To compute a reward bonus, a predictor network $f(s^+)$ is trained to predict the outputs of a fixed, randomly initialised prior network $g(s^+)$, on all hyper-states $s^+$ visited by the agent so far in meta-training.
The mismatch between those predictions is low for frequently visited hyper-states and high for novel hyper-states. Formally we define the reward bonus for a hyper-state $s^+_t=(s_t, b_t)$ as
\begin{equation} \label{eq:bonus:hyper}
r^\text{hyper}(s^+_t) = || f(s^+_t) - g(s^+_t) ||^2 .
\end{equation}
We parameterise the predictor network $f_\omega$ with $\omega$ and train it alongside the policy and VAE.

\textbf{Approximate Hyper-State Exploration.}
To meta-learn an (approximately) Bayes-optimal policy, we need access to the belief over tasks at every timestep $t$ during which the policy interacts with the environment. 
To this end, we use VariBAD \citep{zintgraf2020varibad}, because it provides a belief representation using a single vector.
VariBAD trains an inference procedure using a VAE alongside the policy to obtain approximate beliefs, which are represented by the mean and variance of a latent Gaussian distribution.

At the beginning of meta-training, the beliefs do not sufficiently capture task information.
If the policy does not explore and always gets a sparse reward which is uninformative w.r.t. the task, we fail to meta-learn to perform belief inference.
The policy should therefore seek states where the VAE is not yet trained well. 
As a proxy for this, we use the VAE reconstruction error for the reward and states at the current timestep as a reward bonus: 

\begin{align} \label{eq:bonus:error}
	r^\text{error}(r_t, s_t) = -
	\mathbb{E}&_{q_\phi(m|\tau_{:t})} 
	\Big[ 
	\log p_\theta (r_t | s_{t-1}, a_{t-1}, s_t, m) \nonumber \\
	& \hspace{0.6cm} + \log p_\theta (s_t | s_{t-1}, a_{t-1}, m) 
	\Big] .\end{align}

Since $r_t$ and $s_t$ were observed in $\tau_{:t}$, the encoder $q$ has all data to encode the information needed by the decoder $p$ to predict the current reward and state transition. 
Early in training, these predictions are inaccurate in states where the rewards/transitions differ a lot across tasks. 
Therefore this exploration bonus incentivises the agent to visit states that provide crucial training data for the VAE. 
In practice, this reward bonus is computed using one Monte Carlo sample from $q$.
If only one aspect (reward or transitions) change across tasks, VariBAD only learns the respective decoder, and so we only use the respective reward bonus.  

\textbf{Meta-Training Objective.}
Putting these bonuses together, the new objective for the agent is
\begin{align} 
\hat{\mathcal{J}}^+(\psi) = 
& ~ \mathbb{E}_{b_0, T^+, \pi_\psi} 
\Big[ 
\sum_{t=0}^{H^+-1} 
\gamma^t R^+(r_{t+1} | s^+_t, a_t, s^+_{t+1}) \nonumber \\
& + \lambda_h r^\text{hyper}(s^+_{t+1}) + \lambda_e r^\text{error}(r_{t+1}, s_{t+1})
\Big] .
\label{eq:bonus_objective}
\end{align}
While in principle these bonuses tend towards zero during meta-training, we anneal their weights ($\lambda_h$, $\lambda_e$) over time. 
This prevents the policy to keep meta-exploring at meta-test time and ensure that it maximises only the expected online return.
Algorithm \ref{pseudo_code} shows pseudo-code for HyperX.
Implementation details are given in Appendix \ref{appendix:implementation_details}.

\begin{algorithm}[t]
	\caption{HyperX Pseudo-Code} 
	\label{pseudo_code}
	\begin{algorithmic}
		\STATE {\bfseries Input:}  Distribution over MDPs $p(M)$
		\STATE {\bfseries Initialise:} Encoder $q_\phi$, decoder $p_\theta$, policy $\pi_\psi$, RND predictor network $f_\omega$, buffer $\mathcal{B}=\{s_0, b_0\}$
		\FOR{$k=1,\dotsc,K$}
		\STATE{Sample environments $\mathbf{M}=\{M_i\}_{i=1}^N$ where $M_i\sim p$ }
		\FOR{$M_i\in\mathbf{M}$}
		\STATE{Reset $s_0, h_0, b_0$ }
		\FOR{$t=0,\dotsc,T-1$}
		\STATE{Choose action: $a_{t}=\pi_\psi(s_{t}, b_{t})$}
		\STATE{Step environment:  $s_{t+1}, r_{t+1} = M_i.step(a_{t})$}
		\STATE{Update belief: $b_{t+1} = q_\phi(s_{t+1}, a_t, r_{t+1}, h_t)$}
		\STATE{Compute exploration bonuses:\\ 
			\hspace{0.5cm} $r^\text{hyper}(s_{t+1}^+)$ using Eq \eqref{eq:bonus:hyper} \\
			\hspace{0.5cm} $r^\text{error}(r_{t+1}, s_{t+1})$ using Eq \eqref{eq:bonus:error}}
		\STATE{Add data to buffer: \\ 
			\hspace{0.5cm} $\mathcal{B}^p.add(s_{t+1}, b_{t+1}, a_t, r_{t+1}, r_{t+1}^\text{hyper}, r_{t+1}^\text{error})$\label{alg:line:policy_buffer}}
		\ENDFOR
		\ENDFOR
		\STATE{Update VAE, policy, and RND predictor network:}
		\STATE{\hspace{0.1cm}  $(\phi, \theta) \leftarrow (\phi, \theta) + \alpha_{(\phi, \theta)} ~\nabla_{(\phi, \theta)} \sum_{t=0}^{H+} ELBO_t(\phi, \theta)$}
		\STATE{\hspace{0.1cm} $\psi \leftarrow \psi + \alpha_\psi ~\nabla_\psi \hat{\mathcal{J}}(\psi)$ using Eq \eqref{eq:bonus_objective} \label{alg:line:policy_update} }
		\STATE{\hspace{0.1cm} $\omega\leftarrow \omega - \alpha_\omega ~ \nabla_\omega \mathbb{E}_{s^+\sim \mathcal{B}} \left[ ~|| f_\omega(s^+) - g(s^+)||^2_2~ \right] $ }
		\ENDFOR
	\end{algorithmic}
\end{algorithm}


\section{Related Work} \label{sec:related}

\textbf{Exploration Bonuses.}
Deep RL has been successful on many tasks, and naive exploration via sampling from a stochastic policy is often sufficient if rewards are dense.
For hard exploration tasks this performs poorly, and a variety of more sophisticated exploration methods have been proposed.
Many of these reward novel states, often using count-based approaches to measure novelty \citep{strehl2008analysis, bellemare2016unifying, ostrovski2017count, tang2017exploration}.
A prominent method is Random Network Distillation (RND) for state-space exploration in RL \citep{osband2017deep, osband2018randomized, burda2019exploration, ciosek2020conservative}.
We use it for \emph{hyper}-states in this paper.
We further use an exploration bonus based on the VAE reconstruction error of rewards and transitions. 
Prediction errors of environment dynamics are used to explore the MDP state space, by, e.g.,  \citet{achiam2017surprise, burda2018large, pathak2017curiosity, schmidhuber1991possibility, stadie2015incentivizing}.

\textbf{Meta-Learning Task-Exploration.}
Meta-learning how to adapt quickly to a new tasks often includes learning efficient task-exploration, i.e., how to explore an unknown environment.
We distinguish \emph{few-episode learning} where the agent has several episodes for exploration and maximises final episodic return, and \emph{online adaptation} where performance counts from the first timestep in a new environment, and the agent has to carefully trade off exploration and exploitation.

A popular approach to few-episode learning is gradient-based meta-learning.
Here the agent collects data, then performs a gradient update, and is evaluated afterwards.
Examples are MAML \citep{finn2017model} and follow-up work addressing task-exploration \citep{rothfuss2018promp,stadie2018some}; 
learning separate exploration and exploitation policies \citep{gurumurthy2019mame, liu2020explore, zhang2020learn}; 
or doing task-exploration based on sampling \citep{gupta2018meta} where in particular PEARL \citep{rakelly2019efficient} exhibits behaviour akin to posterior sampling. 
Few-episode learning methods maximise episodic return, and can often not be Bayes-optimal by design.

In this paper we consider the online adaptation setting where we instead want to maximise \emph{online} return.
While computing the exact solution -- the Bayes-optimal policy -- is infeasible for all but the simplest environments, approximate solutions can be meta-learned \citep{ortega2019meta, mikulik2020meta}. 
When using recurrent policies that receive rewards and actions as inputs in addition to the states \citep{duan2016rl, wang2016learning}, learning how to explore happens within the policy network's dynamics, and can be seen as implicitly maintaining a task belief.
\citet{humplik2019meta} and \citet{zintgraf2020varibad} develop methods that represent this belief explicitly, by meta-learning to perform inference either using privileged task information during training  \citep{humplik2019meta}, or by meta-learning to perform inference in an unsupervised way \citep{zintgraf2020varibad}.
To the best of our knowledge, all existing meta-learning methods for online adaptation rely on myopic exploration during meta-training. 
As we observe empirically (Sec \ref{sec:experiments}), this can cause them to break down if rewards are too sparse.

\textbf{Meta-Exploration.}
Two recent works also study the problem of exploration \emph{during} meta-training, albeit for few-episode learning.
This setting can be more forgiving when it comes to the task-exploration behaviour since the agent has multiple rollouts to collect data, and is reset to the starting position afterwards. 
Still, similar considerations about meta-exploration apply.
\citet{zhang2020learn} propose MetaCURE, which meta-learns a separate exploration policy that is intrinsically motivated by an exploration bonus that rewards information gain.
\citet{liu2020explore} propose DREAM, where a separate exploration policy is trained to collect data from which a task embedding (pre-trained via supervision with privileged information) can be recovered.
These methods can, in principle, still suffer from poor meta-exploration if rewards are so sparse that there is no signal to begin with, and information gain / task embedding recovery cannot be measured. 
We empirically compare to MetaCURE and find that this is indeed true (Sec \ref{sec:experiments:cheetah}). 
On the challenging sparse ML1 Meta-World tasks \cite{yu2019meta}, HyperX outperforms MetaCURE by a large margin even though MetaCURE receives more time to explore (Appendix \ref{appendix:additional_results:metaworld}).

If available, privileged information can be used during meta-training to guide exploration, such as expert trajectories \citep{dorfman2020offline}, dense rewards for meta-training but not testing \citep{rakelly2019efficient}, or ground-truth task IDs / descriptions \citep{liu2020explore, kamienny2020learning}. 
HyperX works well even if such information is not available.

\textbf{Exploration in POMDPs.}
Meta-exploration is related to exploration when learning in partially observable MDPs (POMDPs, \citet{cassandra1994acting}), of which BAMDPs are a special case.
This topic is mostly studied on small environments.
Similar to our work, \citet{cai2009learning} incentivize exploration in under-explored regions of belief
space.
However, they use two separate
policies for exploration and exploitation and rely on Bayesian learning to update them, restricting this to small discrete state spaces.
Several authors \citep{poupart2008model,ross2008bayes,doshi2008reinforcement,ross2011bayesian} explore model-based Bayesian reinforcement learning in partially observable domains. 
By relying on approximate value iteration to solve the planning problem, they are also restricted to small environments.
To our knowledge, only \citet{yordan2019intrinsic} provides some initial results on a
simple environment using Random Network Distillation.
They propose various ways to deal with the non-stationarity of the latent embedding such as 
using a random \emph{recurrent} network that aggregates past trajectories.


\section{Empirical Results} \label{sec:experiments}

We present four experiments that illustrate how and why HyperX helps agents meta-learn good online adaptation strategies (Sec \ref{sec:experiments:treasure}-\ref{sec:experiments:cheetah}), and results on sparse MuJoCo Ant-Goal to demonstrate that HyperX scales well (Sec \ref{sec:experiments:ant_goal}).

Many standard meta-RL benchmarks do not require much exploration, in the sense that there is no room for improvement via better exploration, such as the $2$D navigation Pointrobot (Appendix \ref{appendix:additional_results:point_robot}), Meta-World ML1 even with sparse rewards (Appendix \ref{appendix:additional_results:metaworld}), or the otherwise challenging dense AntGoal environment (Appendix \ref{appendix:antgoal_results}).

\subsection{Treasure Mountain} \label{sec:experiments:treasure}

\begin{figure}[t]
	\centering
	\begin{subfigure}{0.96\linewidth}
		\centering
		\includegraphics[width=\linewidth]{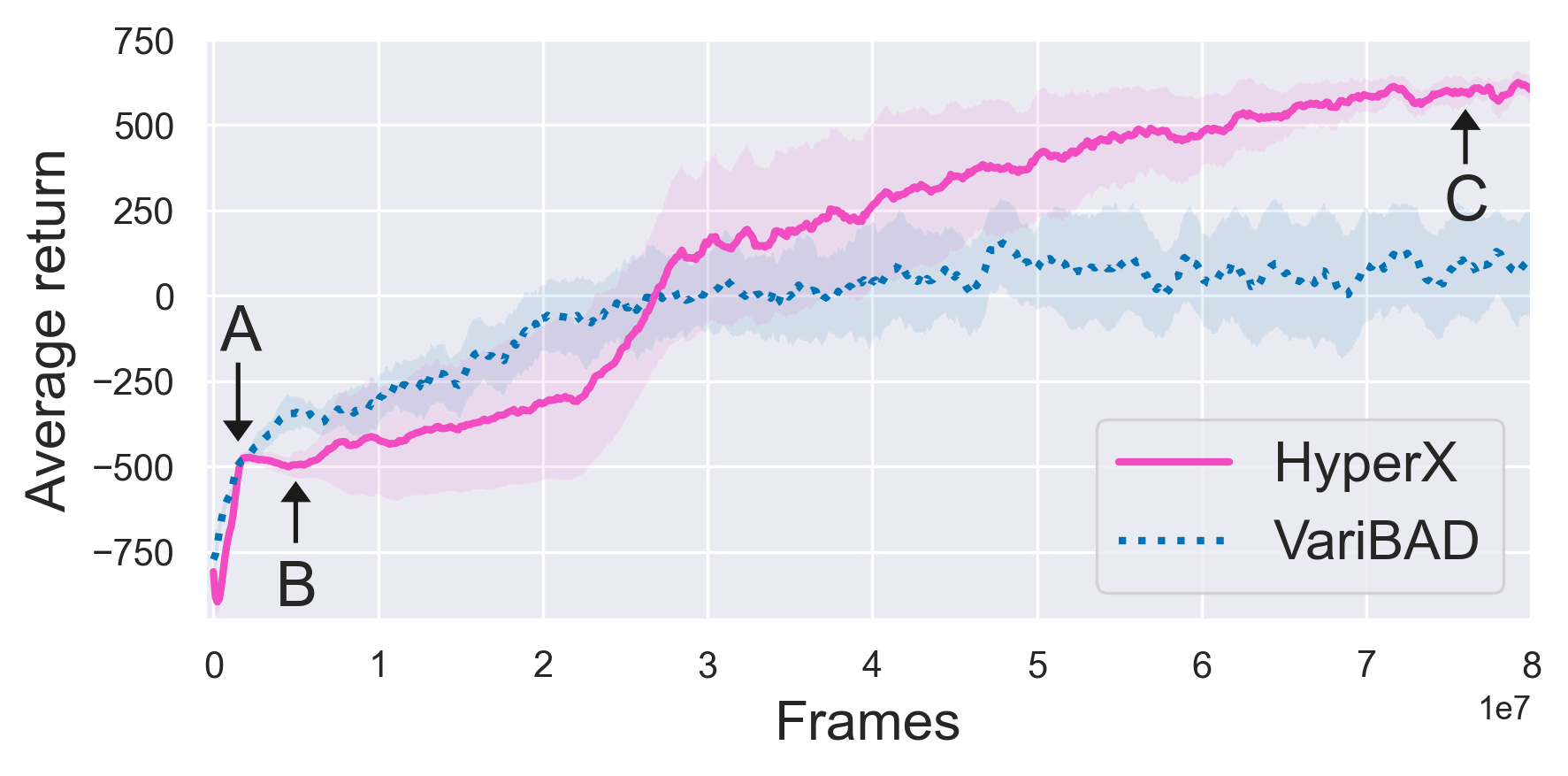}
	\end{subfigure}
	\\
	\begin{subfigure}{0.3\linewidth}
		\centering
		\includegraphics[width=\linewidth]{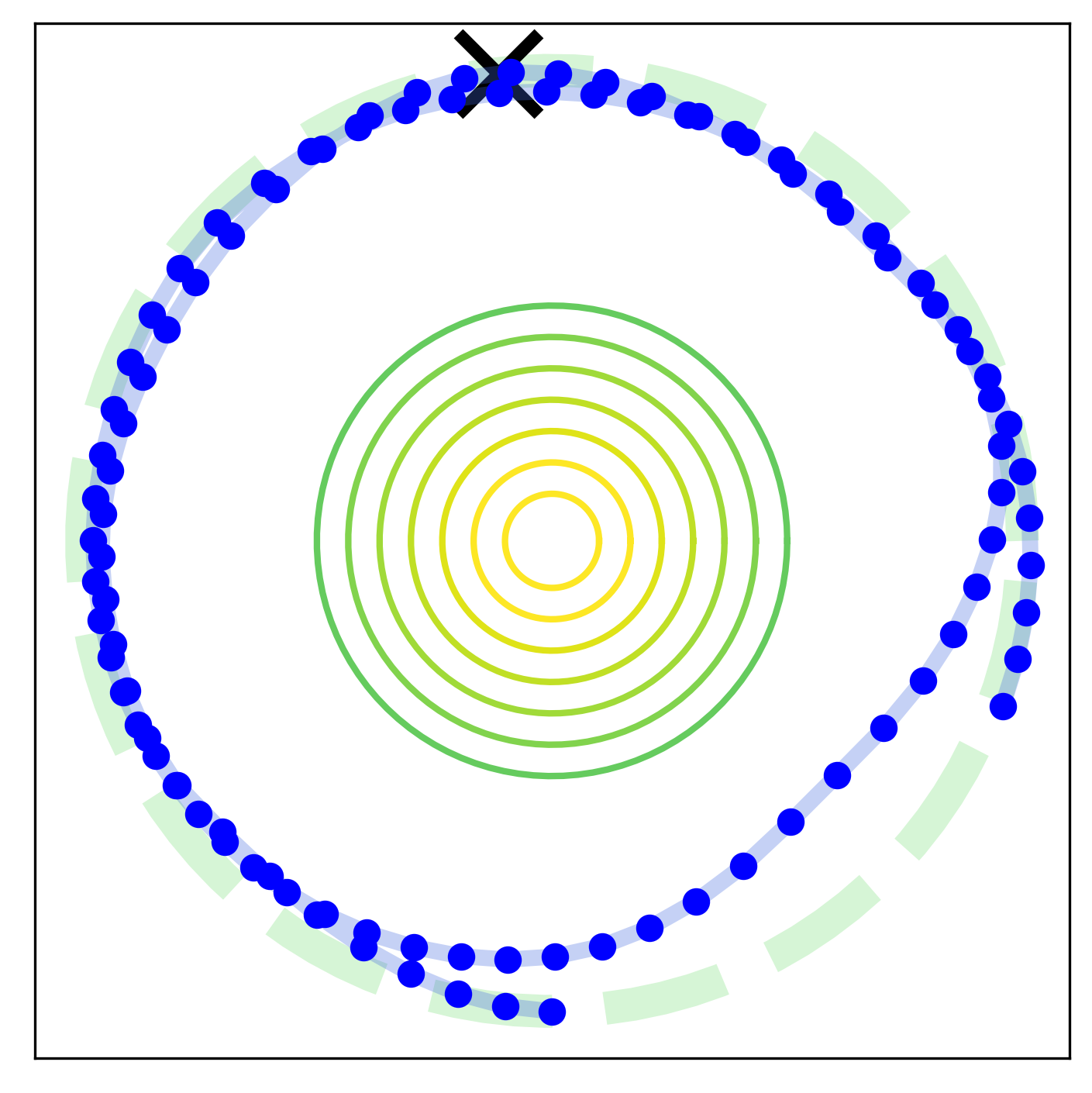}
		\caption{}
		\label{fig:treasure_A}
	\end{subfigure}%
	\begin{subfigure}{0.3\linewidth}
		\centering
		\includegraphics[width=\columnwidth]{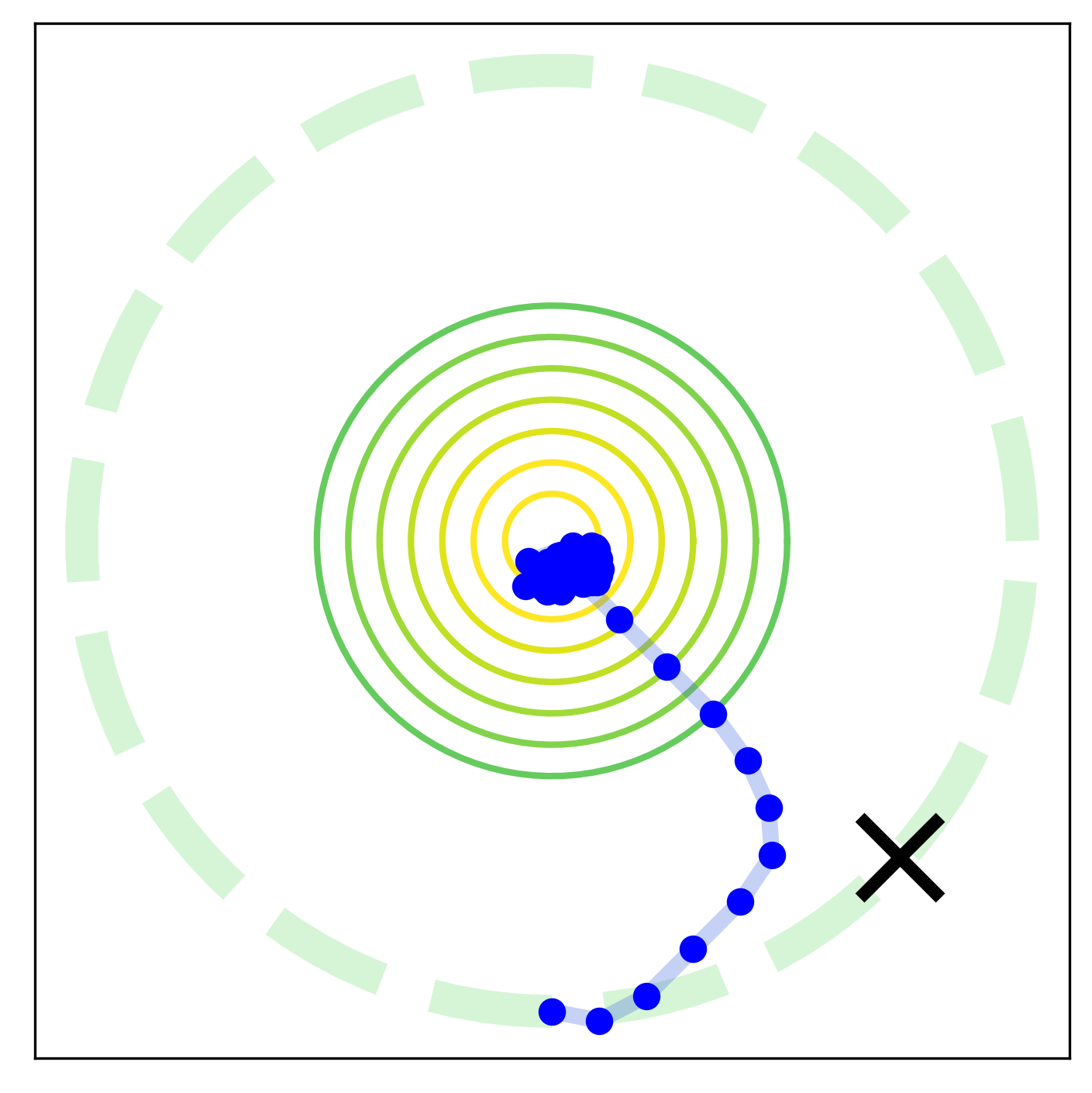}
		\caption{}
		\label{fig:treasure_B}
	\end{subfigure}
	\begin{subfigure}{0.3\linewidth}
		\centering
		\includegraphics[width=\columnwidth]{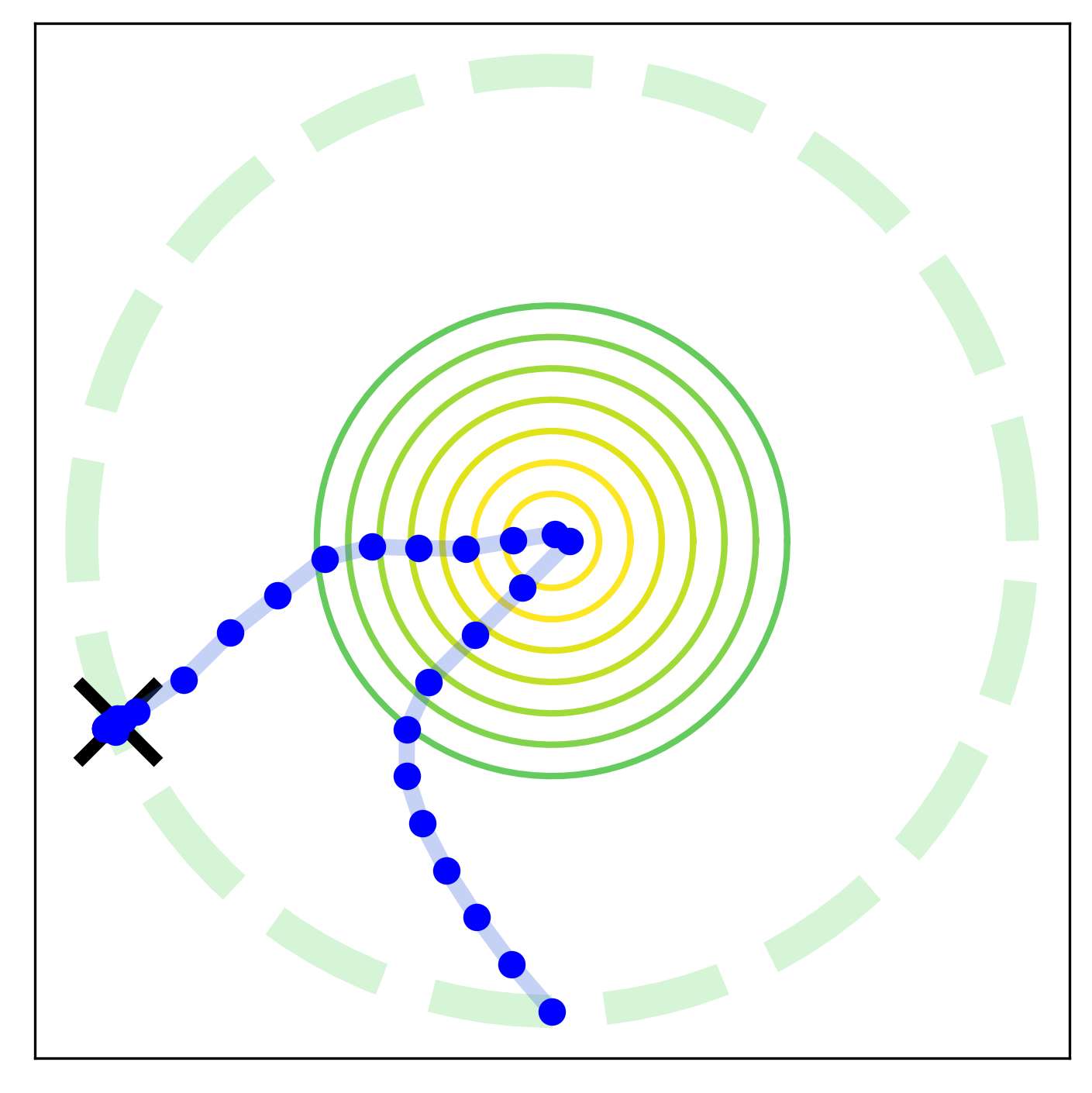}
		\caption{}
		\label{fig:treasure_C}
	\end{subfigure}
	\caption{
		\textbf{Treasure Mountain Results.}
		Top: Learning curves for HyperX and VariBAD ($10$ seeds, $95$\% confidence intervals shaded).
		Bottom: Behaviour of the HyperX agent at different stages of training.
		HyperX learns the superior task-exploration strategy of climbing the mountain to see the treasure, and going there directly after. 
		VariBAD learns the inferior strategy of walking around the circle until finding the treasure (see rollouts in Appendix \ref{appendix:additional_results:treasure_mountain}).
	}
	\label{fig:treasure_rollouts}
\end{figure}

We consider our earlier example (Fig \ref{fig:comic_treasure}),
where the agent's task is  to find a treasure  hidden in tall grass. 
There are two good task-exploration strategies: 
(s1) search the grass until the treasure is found, 
or 
(s2) 
climb the mountain, spot the treasure, and go there directly. 
The latter strategy has higher expected return, 
but is harder to meta-learn since 
(a) climbing the mountain is discouraged by negative rewards and 
(b) the agent must meta-learn to interpret and remember the treasure location it sees from the mountain. 
 
We implement this as follows (details in Appendix \ref{appendix:environments:mountain}): 
the treasure can be anywhere along a circle. 
The agent gets a sparse reward when it reaches the treasure, and a time penalty otherwise.
Within the circle is the mountain, represented by a smaller circle. 
Walking on it incurs a higher time penalty.
The agent's observation is $4$D: its $x$-$y$ position, and the treasure's $x$-$y$ coordinates, which are \emph{only} visible from the mountain top.
The agent starts at the bottom of the circle and has one rollout of $100$ steps to find the treasure. 

Figure \ref{fig:treasure_rollouts} shows the learning curves of VariBAD (which uses no exploration bonuses for meta-training) and HyperX, with HyperX performing significantly better.
Figures \ref{fig:treasure_A}-\ref{fig:treasure_C} show the behaviour of HyperX at different times during training. 
At the beginning (\ref{fig:treasure_A}), it explores along the circle (but does not stop at the treasure and explores further) and its performance increases. 
Then, it discovers the mountain top: because the VAE reconstruction error $r^{error}$ is high there, it initially just stays there (\ref{fig:treasure_B}).
Performance drops since the penalty for climbing the mountain is slightly higher than the time penalty the agent gets otherwise.
Finally, at the end of training (\ref{fig:treasure_C}) it learns the optimal strategy s2 (consistently across all $10$ seeds).
Inspection of the rollouts show that VariBAD and other methods for online adaptation (RL$^2$ \citep{duan2016rl, wang2016learning} and Belief Learning \citep{humplik2019meta}) always only meta-learn the inferior task-exploration strategy s1 (see Appendix \ref{appendix:additional_results:treasure_mountain}).

When using \emph{only} $r^{hyper}$, the agent only learns the inferior strategy s1: early in training, hyper-states are meaningless and the agent stops exploring the mountain top.
When using only $r^{error}$, the agent learns the superior strategy s2 around $70\%$ of the time.
For learning curves see Appendix \ref{appendix:additional_results:treasure_mountain}.

This experiment shows that HyperX tries out different task-exploration strategies during meta-training, and can therefore meta-learn a superior exploration strategy even when it is expensive in the short term, but pays off in the longer run.

\subsection{Multi-Stage Gridworld} \label{sec:experiments:gridworld}

\begin{figure}
	\centering
	\includegraphics[width=0.68\linewidth]{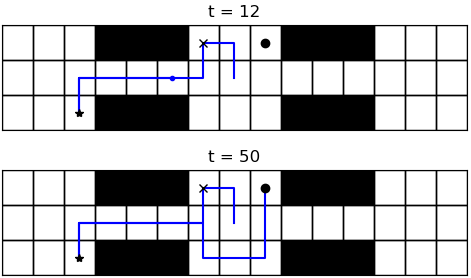}
	\caption{
		\textbf{Multi-Stage Gridworld.} 
		Goal 1 ($\times$) unlocks goal 2 ($*$), which unlocks goal 3 ($\bullet$). Example behaviour of HyperX in blue.}
	\label{fig:room}
\end{figure}
\begin{figure}
	\centering
	\includegraphics[width=\linewidth]{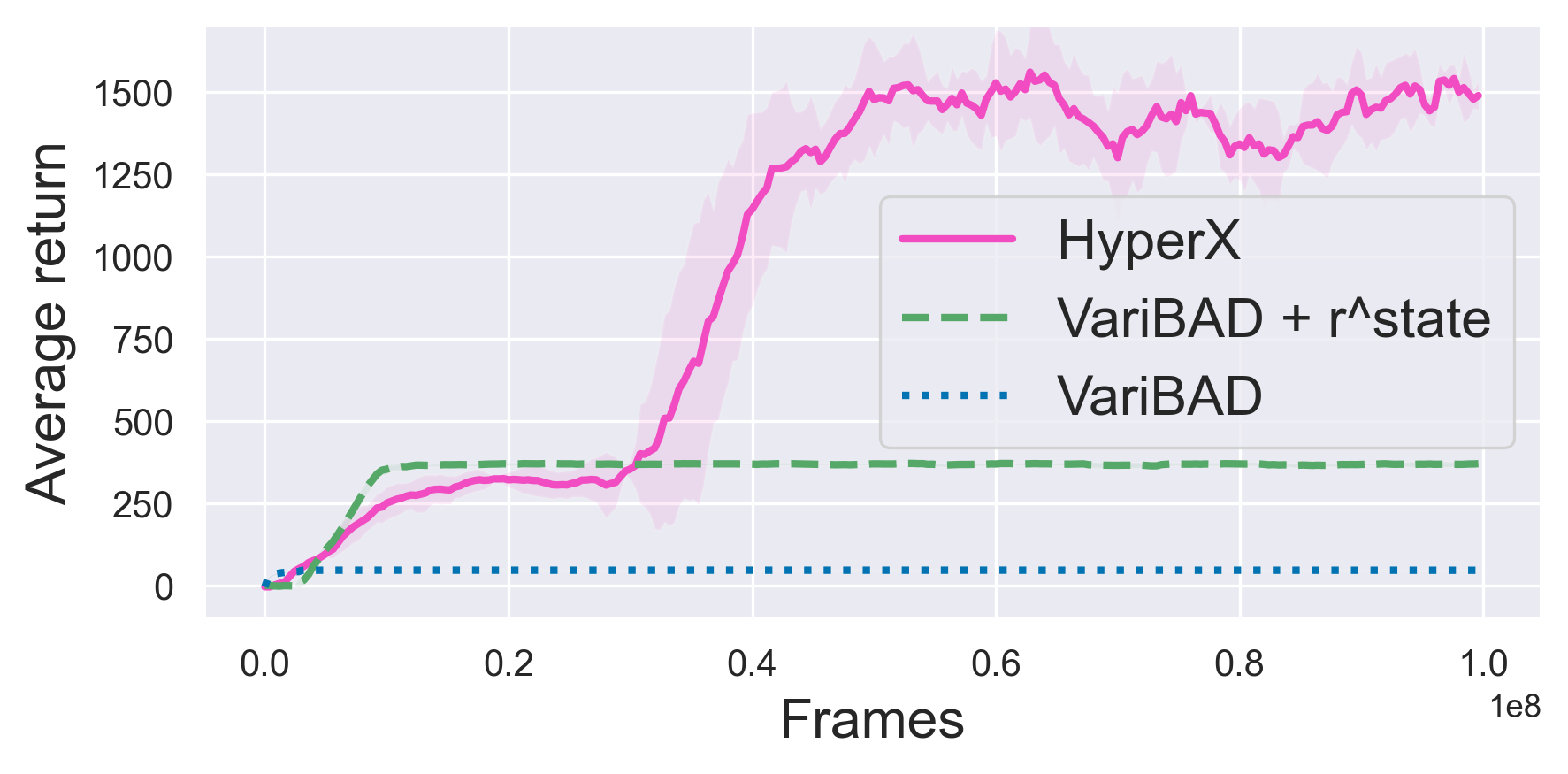}
	\caption{\textbf{Multi-Stage Gridworld Learning Curves.} ($3$ seeds)}
	\label{fig:performance}
\end{figure}

Next, we consider a partially observable multi-stage gridworld which illustrates how,
without the appropriate exploration bonuses on hyper-states, 
existing methods can converge prematurely to a local optimum.

The gridworld is illustrated in Figure \ref{fig:room}:
three rooms are connected by narrow corridors, and three (initially unknown) goals (G1-G3) are placed in corners of rooms:
The goals provide increasing rewards, i.e. $r_{1}=1$, $r_{2}=10$ and $r_{3}=100$, but are only sequentially unlocked; G2 ($r_{2}$) is only available after G1 has been reached; G3 ($r_{3}$) is only available after G2 has been reached.
The environment is partially observable \citep{poupart2008model, cai2009learning} as the agent only observes its position in the environment and not which goals are unlocked.
If the agent is not on an (available) goal it gets $r=-0.1$. 
G1 and G3 are always in the middle room, G2 always in an outer room on the same side as G1.
The agent starts in the center of the middle room and has $H=50$ steps. The best strategy is to search the first room for G1, then search the appropriate room for G2, and then return to the middle room to find G3.

Figure \ref{fig:performance} compares VariBAD, VariBAD with state-novelty bonus, and HyperX.
VariBAD learns to reach G1 and remains there, effectively receiving only $r_1$ at every timestep.
VariBAD$+r(s)$ learns to find G2 and stay there, but fails to find G3. 
Only HyperX solves the problem (see behaviour in Figure \ref{fig:room}).
Methods which use a purely state-based exploration bonus such as VariBAD$+r(s)$ are unable to find G3 in the middle room as those states $s$ (not hyper-states $(s,h)$) appear already sufficiently explored.
In contrast, a novelty bonus on the hyper-state $r(s,h)$ like in HyperX leads to a high novelty bonus in the middle room once G2 is found because the belief changes.

These results show that without the right exploration bonuses during meta-training, the agent can prematurely converge to a suboptimal solution. 
Additionally, we see that that HyperX can handle this degree of partial observability.

\subsection{Sparse HalfCheetahDir} \label{sec:experiments:cheetah}

\begin{table}[h!]
	\centering
	\subfloat[Method Comparison]{%
		\hspace{.5cm}%
		\begin{tabular}[t]{lcr}
			\toprule
			Method & Avg Return \\
			\midrule
			VariBAD & $-1.1$ \\
			E-MAML & $-0.4$ \\
			ProMP & $-0.4$ \\
			Humplik et al. & $-0.1$ \\
			RL$^2$ & $-0.7$ \\
			PEARL & $-0.1$ \\
			\midrule
			HyperX & $819.6$ \\
			\bottomrule
		\end{tabular}%
		\label{table:cheetah_methods}
		\hspace{.5cm}%
	}\hspace{1cm}
	\subfloat[Ground Truth Beliefs + Exploration Bonuses]{%
		\hspace{.5cm}%
		\begin{tabular}[t]{lc}
			\toprule
			Method & Avg Return \\
			\toprule
			Belief Oracle & $-3.0$ \\
			Belief Oracle + $r(b)$ & $-3.6$ \\
			Belief Oracle + $r(s)$ & $639$  \\
			Belief Oracle + $r^\text{hyper}$ & $824$ \\
			\bottomrule
		\end{tabular}%
		\label{table:cheetah_belieforacle}
		\hspace{.5cm}%
	}\hspace{1cm}
	\subfloat[Ablation Studies.]{%
	\hspace{.5cm}%
	\begin{tabular}[t]{lc}
		\toprule
		Method & Avg Return \\
		\toprule
		HyperX , $r^\text{error}$ only & $-0.7$ \\
		HyperX , $r^\text{hyper}$ only & $462$ \\
		\midrule
		RL$^2$ + $r^\text{state}$ & 463 \\
		RL$^2$ + $r^\text{hyper}$ & 477 \\
		Humplik et al. + $r^\text{hyper}$ & 840 \\
		Humplik et al. + $r^\text{hyper}$ + $r^\text{error}$ & $850$ \\
		\midrule
		VariBAD + $r^\text{metaCURE}$ & -0.3 \\
		VariBAD + $r^\text{metaCURE}$ + $r^\text{state}$ & 548 \\
		VariBAD + $r^\text{metaCURE}$ + $r^\text{hyper}$ & 811 \\
		\bottomrule
	\end{tabular}%
	\label{table:cheetah_ablations}%
	}%
	\caption{\textbf{HalfCheetahDir Meta-Test Performance.} 
		(a) 
		HyperX successfully solves this task, while existing meta-learning methods fail.
		(b) 
		Not even an agent with access to the correct belief is able to solve this task without appropriate exploration bonus.
		(c) 
		Both exploration bonuses are necessary to succeed, but different realisations with similar effects (e.g., metaCURE) work as well. 
		The bonuses can be used with other methods, if they provide an approximate belief (for $r^{hyper}$) and a way of measuring how good the inference is (for $r^\text{error}$).
	}
	\label{table:cheetah_results}
\end{table}

\begin{figure}[h]
	\centering
	\begin{subfigure}{0.95\linewidth}
		\centering
		\includegraphics[width=\linewidth]{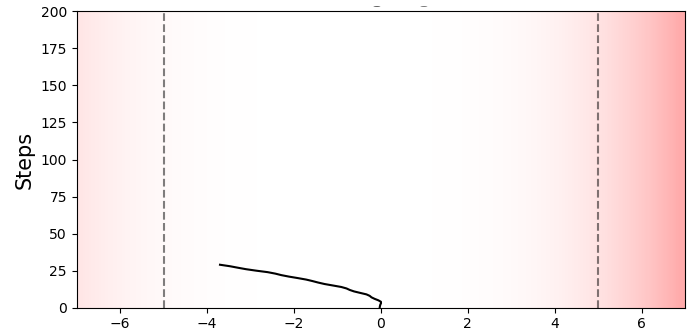}
	\end{subfigure}
	\begin{subfigure}{0.95\linewidth}
		\centering
		\includegraphics[width=\linewidth]{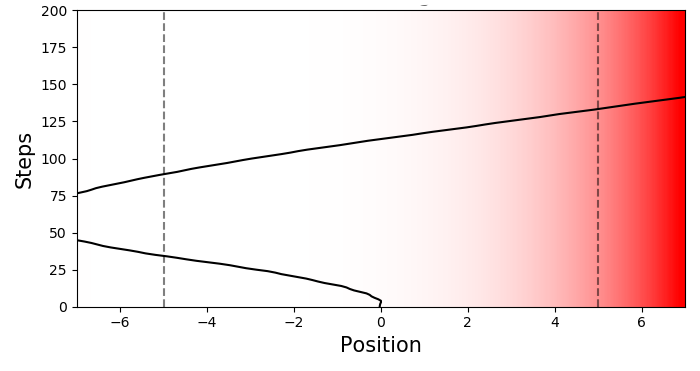}
	\end{subfigure}%
	\caption{
		\textbf{HalfCheetahDir Example Rollouts} for the Belief Oracle, early in meta-training ($1e6$ frames), trained with the hyper-state-bonus $r^{hyper}(s^+)$.
		The $y$-axis denotes time in agent steps. 
		The background visualises the hyper-state-bonus: darker means higher bonus. 
		\textbf{Top}: At the beginning of the episode, the agent's belief is the prior, and the exploration bonus incentivises it to explore away from the familiar start position. 
		\textbf{Bottom}: Once the agent enters the dense-reward zone, it infers the task and updates its belief.
		Now, the states on the right side seem novel since the agent has not seen them together with the posterior belief.
	}
	\label{fig:reward_bonus_cheetah}
\end{figure}

To demonstrate the effect of the different exploration bonuses, we consider the following example for which we can compute exact beliefs.
The environment is based on the HalfCheetah-Dir MuJoCo environment, which is commonly used in meta-RL \citep[e.g.,][]{finn2017model, rakelly2019efficient}.
The prior distribution $p(M)$ is uniform over the two tasks ``walk forward'' and ``walk backward''. 
In the dense version the agent is rewarded according to its ($1$D) velocity in the correct direction. 
We consider a \emph{sparse} version without resets: the agent only receives the dense reward once it walks sufficiently far away from its starting position, outside an interval $[-5, 5]$ (and a control penalty otherwise), and has $200$ environment steps to adapt.
This makes it much more difficult to find the optimal adaptation strategy, which is to walk far enough in one direction to infer the task, and turn around in case the direction was wrong. 

Without dense rewards, existing meta-learning algorithms fail to learn this strategy, 
as is the case for RL$^2$ \citep{duan2016rl, wang2016learning}, PEARL \citep{rakelly2019efficient}, ProMP \citep{rothfuss2018promp}, E-MAML\footnote{E-MAML/ProMP/PEARL are not designed to adapt within a single episode, so we do the gradient update (E-MAML/ProMP) / posterior sampling (PEARL) after half an episode.} \citep{stadie2018some} and VariBAD \citep{zintgraf2020varibad}, as shown in Table \ref{table:cheetah_methods}. 
HyperX in contrast successfully meta-learns the correct task-adaptation strategy.
For all baselines, we used the available open source code.

\textbf{Exploration in Exact Hyper-State Space.}
To investigate how the exploration bonuses in HyperX help solve this task, we first assume that we have access to the true hyper-state $s^+_t {=}(s_t, b_t)$, including the true belief which we define as follows.
The prior belief is $b_0{=}[0.5, 0.5]$ and it can be updated to the posterior belief $b{=}[1, 0]$ (left) or $b{=}[0, 1]$ (right) once the agent observes a single reward outside of the interval $[-5, 5]$. 
Since we can manually compute this belief, we can train a Belief Oracle using standard reinforcement learning, by conditioning the policy on the exact hyper-state.
Table \ref{table:cheetah_belieforacle} shows the performance of the Belief Oracle, with and without reward bonuses.
Without the bonus, even this Belief Oracle does not learn the correct
behaviour for this seemingly simple task.
When adding the exploration bonus $r^\text{hyper}(b, s)$ on the hyper-state, the policy learns approximately Bayes-optimal behaviour. 

Figure \ref{fig:reward_bonus_cheetah} shows how the bonuses incentivise the agent to explore, with the red gradient in the background visualising the reward bonus (darker meaning more bonus).
When the agent walks outside the sparse-reward interval and updates its belief, the reward bonus in the opposite direction becomes high since it has not yet visited that area with the updated belief very often. 
Table \ref{table:cheetah_results} (top) shows that a policy trained with a reward bonus only on the state, $r(s)$, performs worse. 
The reason is that the agent is not incentivised to explore states to the far right after its belief has changed.
Inspection of the learned policies shows that agents trained with a state exploration bonus do go outside the interval, and just return to and stay in the sparse-reward zone  if the direction was wrong (see Appendix \ref{appendix:cheetah_results}).

\textbf{HyperX: Exploration in Approximate Hyper-State Space.}
Above we assumed access to the true belief $b_t$.
When meta-learning how to perform approximate belief inference alongside the policy however, these beliefs change over time and are initially inaccurate.
As Table \ref{table:cheetah_ablations} (top) shows, using only the hyper-state exploration bonus $r^\text{hyper}$, which worked well for the Belief Oracle, performs sub-optimally. 
This is because early in training the belief inference is inaccurate, and the hyper-state bonus is meaningless: the agent prematurely and wrongly assumes it has sufficiently explored.
Only when adding the error reward bonus $r^\text{error}$ as well to incentivise the agent to explore areas where the belief inference makes mistakes, can we meta-learn approximately Bayes-optimal behaviour for this task.
Using only the error reward bonus $r^\text{error}$ performs poorly as well.

\textbf{Different meta-learners.}
While we build HyperX on VariBAD, the same exploration bonuses can be used for other meta-learning methods that provide (a) a belief representation, and (b) a measure of how good the belief inference is.
One such method is the work by \citet{humplik2019meta} who train a belief model using the ground-truth task description.
This makes meta-learning inference easier, and the exploration bonus $r^\text{error}$ may not be necessary:
for sparse HalfCheetahDir, using only the hyper-state bonus ($r^\text{hyper}$) is sufficient, as shown in Table  \ref{table:cheetah_ablations} (middle).
This result is not directly comparable to HyperX since it uses privileged information, whereas HyperX meta-learns inference in an unsupervised way.
For the method RL$^2$, the RNN hidden state can be used as a belief proxy to compute the hyper-state bonus. 
The second exploration bonus ($r^\text{error}$) however, cannot be estimated because the hidden state is only used implicitly by the agent.
As Table \ref{table:cheetah_ablations} shows, an exploration bonus on the state ($r^\text{state}$) or on the hyper-state ($r^\text{hyper}$) for RL$^2$ is not sufficient to solve the task.

\textbf{Comparison to MetaCURE.}
Recently \citet{zhang2020learn} proposed MetaCURE for meta-exploration. 
They use information gain as an intrinsic reward, defined as the difference between the prediction errors (of states/rewards) given the agent's current experience or given the ground-truth task. 
For the sparse HalfCheetahDir task this is high when the agent first steps over the interval bound. 
Even though MetaCure is defined for episodic task-adaptation, we can use its bonus in the online adaptation setting as well.
Compared to HyperX it requires training two additional prediction networks, and it relies on privileged task information during meta-training to do so.
Table \ref{table:cheetah_ablations} shows that this exploration bonus alone is not sufficient to solve the task -- it only incentivises the agent to go to the interval boundary). 
Adding a hyper-state bonus is required to solve the task.

\subsection{Sparse MuJoCo AntGoal} \label{sec:experiments:ant_goal}

\begin{figure}
	\centering
	\begin{subfigure}{0.6\linewidth}
		\includegraphics[width=\linewidth]{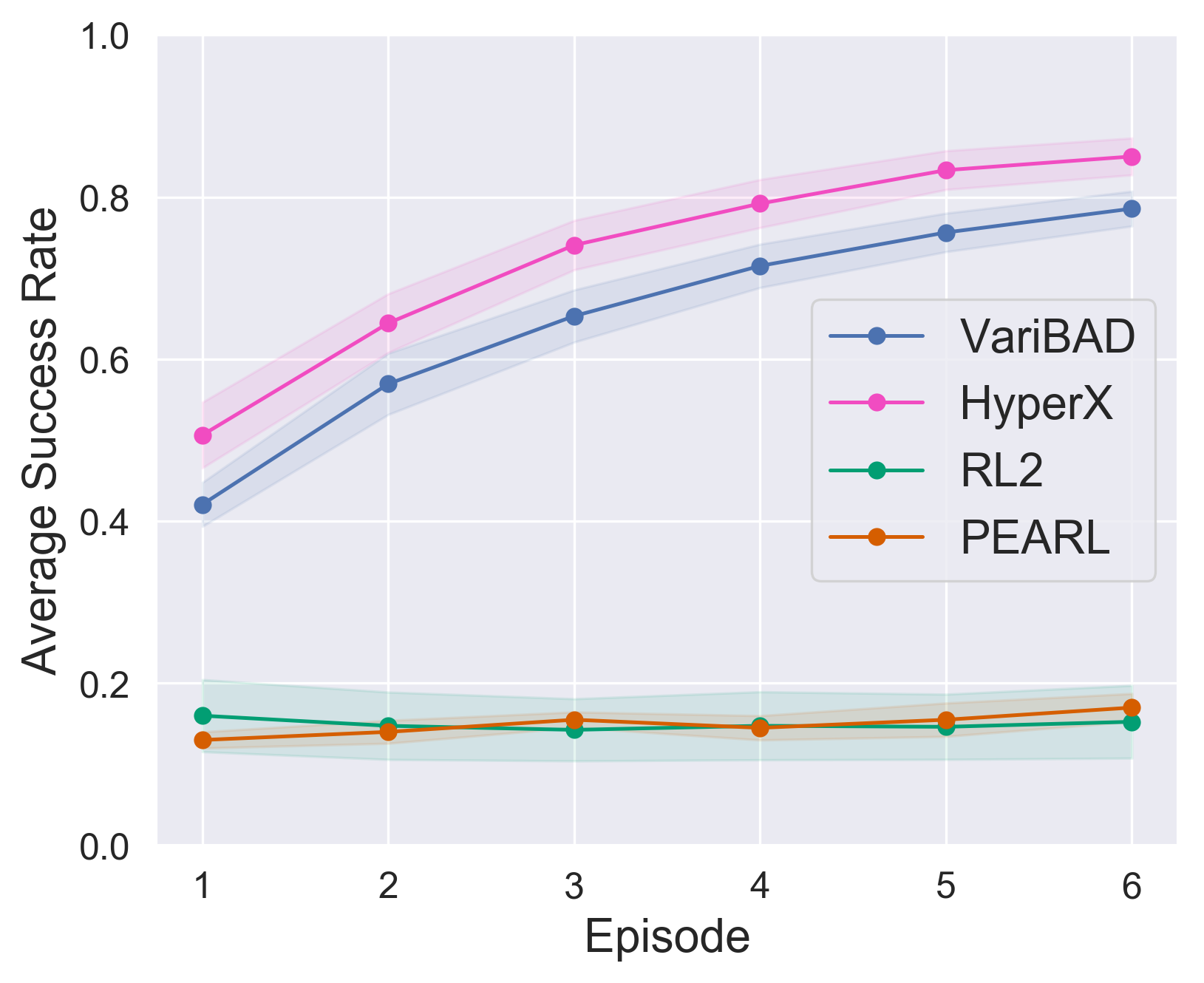}
		\caption{Success Rates}
		\label{fig:antgoal_sparse:success_rate}	
	\end{subfigure}
	\begin{subfigure}{0.35\linewidth}
		\includegraphics[width=\linewidth]{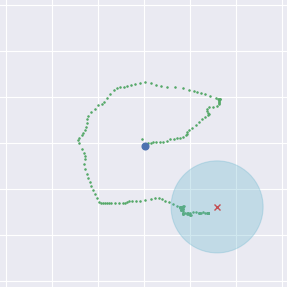}
		\caption{HyperX Rollout}
		\label{fig:antgoal:rollout}
	\end{subfigure}
	\caption{
		\textbf{Sparse AntGoal Results.} 
		\ref{fig:antgoal_sparse:success_rate} shows the success rate per episode ($10$ seeds, standard error shaded). 
		\ref{fig:antgoal:rollout} shows a cherry-picked test rollout of the HyperX agent during the first episode.
	}
\end{figure}

To show that HyperX can scale to more complex environments, we evaluate it on a harder MuJoCo task, a sparse version of the common meta-learning learning benchmark Ant-Goal-2D \citep{rothfuss2018promp, rakelly2019efficient}.
The agent must navigate to an initially unknown goal position. 
In the dense version, the agent gets a reward relative to its distance to the goal.
This version can be solved by current meta-learning results (see Appendix \ref{appendix:additional_results}). 
We make this task significantly harder by sparsifying the rewards, giving the agent the dense reward only if it is within a certain distance of the goal (the blue shaded area in Figure \ref{fig:antgoal:rollout}).
The best exploration strategy is therefore to spiral outwards until a reward signal is found. 

Figure \ref{fig:antgoal_sparse:success_rate} shows the success rate of HyperX, VariBAD, RL$^2$ and PEARL across different episodes, where an agent is successful if it enters the goal circle. 
For RL$^2$, VariBAD, and HyperX, the belief is maintained across episodes by not resetting the RNN (encoder) hidden state.
Both RL$^2$ and PEARL only learn to go to goals close to the starting position, and therefore have low success rate. 
HyperX has a higher success rate than VariBAD. 
This result illustrates that a lack of good exploration can crucially affect final performance: the success rate in the 6th episode is good iff early exploration was good. 
Plots for the returns across episodes and learning curves can be found in Appendix \ref{appendix:antgoal_results}. 

Figures \ref{fig:antgoal:hyperx_rollouts} and \ref{fig:antgoal:varibad_rollouts} (Appendix \ref{appendix:antgoal_results}) show example behaviours of the meta-trained HyperX and VariBAD agent at test time. 
HyperX learns to efficiently search the space of possible goals, occasionally in the shape of a spiral (Fig \ref{fig:antgoal:rollout}), finding the goal in the first episode.
VariBAD can also learn to search in a spiral but is not as efficient and more likely to fail.  
Once the agent reaches dense-reward radius around the goal, it is able to determine where the goal is and heads there directly. 
In subsequent episodes, the agent exploits its knowledge about the goal and returns directly to it.

Overall our empirical results show that HyperX can meta-learn excellent adaptation behaviour on challenging sparse reward tasks where existing methods fail.
We also evaluated HyperX on sparse environments used in the literature, like sparse PointRobot \citep{rakelly2019efficient}, and sparse Meta-World ML1 \citep{yu2019meta}.
However, in both cases, VariBAD can already solve these tasks (Appendix \ref{appendix:additional_results:metaworld}-\ref{appendix:additional_results:point_robot}).


\section{Conclusion}

\textbf{The Meta-Exploration Problem.}
This paper showed that existing meta-learning methods can fail if the environment rewards are not densely informative with respect to the task, and myopic exploration during meta-training is insufficient. 
We highlighted that in this case, special attention needs to be paid to \emph{meta-exploration}.
This applies to many different problem settings, but we focused on online adaptation where the agent aims to maximise expected online return.
Here, task-exploration is particularly challenging since the agent has to trade off exploration and exploitation.

\textbf{Our Solution: HyperX.}
We proposed HyperX, which uses two exploration bonuses to incentivise the agent to explore in approximate hyper-state space during meta-training.
This way, it collects the data necessary to learn approximate belief inference (incentivised by $r^\text{error}$), and tries out different task-exploration strategies during meta-training (incentivised by $r^\text{hyper}$).  
We demonstrated empirically how meta-learning without explicit meta-exploration can fail and why, and showed that HyperX can solve these tasks.

\section*{Software and Data}

Our source code can be found at \url{https://github.com/lmzintgraf/hyperx}.


\section*{Acknowledgements}

We thank Wendelin Böhmer, Tabish Rashid, Matt Smith, Jelena Luketina, Sebastian Schulze, and Joost van Amersfoort for helpful discussions and feedback.
We also thank the anonymous reviewers for their feedback and suggestions that helped improve our paper.
Luisa Zintgraf is supported by the 2017 Microsoft Research PhD Scholarship Program, and the 2020 Microsoft Research EMEA PhD Award.
Maximilian Igl and Cong Lu are supported by the UK EPSRC CDT in Autonomous Intelligent Machines and Systems.
Kristian Hartikainen is funded by the EPSRC.
This work was supported by a generous equipment grant from NVIDIA, 
and enabled in part by computing resources provided by Compute Canada.
This project has received funding from the European Research Council under the European Union's Horizon 2020 research and innovation programme (grant agreement number 637713).

\bibliography{main}
\bibliographystyle{icml2021}


\newpage

\appendix

\begin{center}
	{\textbf{Exploration in Approximate Hyper-State Space \\ for Meta Reinforcement Learning}} \\ \vspace{0.3cm}
	{\centering \Large Supplementary Material} 
\end{center}

\section{Additional Background}

\subsection{Randomised Prior Functions} \label{sec:background:rpfs}

In reinforcement learning, we can use the fact that unseen states can be seen as out-of-distribution data of a model that is trained on all data the agent has seen so far.
Getting uncertainty estimates on states can thus quantify our uncertainty about the value of a state and in turn whether we have explored these states sufficiently. We can think about why exploration purely in the state space $\mathcal{S}$ (which is shared across tasks) is not enough: 
if the agent has explored a state many times in one task and is certain of its value, it should not necessarily exploit this knowledge in a different task, because this same state could have a completely different value.
We cannot view these as separate exploration problems however, since we also have to try out different deployed exploration strategies and combine the information to meta-learn Bayes-optimal behaviour.

Therefore, we want to incentivise the agent to explore in the hyper-state space $\mathcal{S}^+ = \mathcal{S} \times \mathcal{B}$.
Only if an environment state together with a specific belief has been observed sufficiently often to determine its value should the agent trust its value estimate of that belief-state.
This therefore amounts to exploration in a BAMDP state space, which essentially means trying out different exploration strategies in the environments of the training distribution.  
We use Random Network Distillation (RND) \citep{osband2018randomized, burda2019exploration, ciosek2020conservative} to obtain such uncertainty estimates and review them using the formulation of \citet{ciosek2020conservative} in the following.

Assume we are given a set of training data $\mathcal{D}=\{s_i\}_{i=1}^{N}$ of all states the agent has observed.
To get uncertainty estimates, we first fit $B$ predictor networks $g_j(s)$ ($j=1,\dotsc,B$) to a random prior process $f_j(s)$ each (a network with randomly initialised weights, which is fixed and never updated).
We then estimate the uncertainty for a state $s_*$ as
\begin{equation}
\sigma^2(s_*) = \max(0, ~ \sigma^2_\mu(s_*) + \beta v_\sigma(s_*) - \sigma_A^2),
\end{equation}
where $\sigma^2_\mu(s_*) $ is the sample mean of the squared errors between the $B$ predictor networks and the prior processes; $v_\sigma(s_*)$ is the sample variance of the squared error.
The first quantifies our uncertainty, whereas the second quantifies our uncertainty over what our uncertainty is. 
In practice, $B=1$ is typically sufficient and the second term disappears \citep{ciosek2020conservative}.
The term $\sigma_A^2$ is the aleatoric noise inherent in the data which is an irreducible constant. In theory, this can be learned as well and depends on how much information can be extracted about the value of states and actions from the data. In practice, we set this term to $0$.

Given a hyper-state $s^+_t=(s_t, b_t)$, an ensemble of $B$ prior networks $\{ f^i(s^+) \}_{i=1}^B$ and corresponding predictor networks $\{ h^i(s^+) \}_{i=1}^B$, the reward bonus is defined as
\begin{equation}
r_c(s^+_t) = \max(0, ~ \sigma_mu^2(s^+_t) + \beta v_\sigma(s^+_t) - \sigma^2_A)
\end{equation}
where $\sigma_mu^2(s^+_t) $ is the sample mean of the squared error between prior and predictor networks and $v_\sigma(s^+_t)$ is the sample variance of that error.


\section{Additional Results} \label{appendix:additional_results}

In this section we provide additional experimental results. 
The first two sections are additional environments -- in particular sparse environments used in the literature before, but where we found that our baselines already performed very well. 
In addition, we provide more details and results for the experiments in the main paper.

Implementation details, including hyperparameters and environment specifications, are given in Appendix \ref{appendix:implementation_details}.
The (anonymised) source code is attached as additional supplementary material.

\subsection{Meta-World} \label{appendix:additional_results:metaworld}

\begin{table*}[t]
	\centering
	\setlength{\tabcolsep}{1em}
	\begin{tabular}{l|c|ccc|ccc}
		\multicolumn{2}{c}{} & \multicolumn{3}{c}{Dense Rewards} & \multicolumn{3}{c}{Sparse Rewards} \\ 
		\toprule
		Method & Test Episode & Reach & Push & Pick-Place &  Reach & Push & Pick-Place \\
		\midrule
		MAML$^*$ 	 	& 10 & 48 & 74 & 12 & -  & - & - \\
		PEARL$^*$		& 10 & 38 & 71 & 28 & -  & - & - \\
		RL2$^*$				& 10 & 45 & 87 & 24 & -  & - & - \\
		E-RL2$^+$			& 10 & - & - & - & 28  & 7 & - \\
		MetaCURE$^+$ 	& 10 &  - & - & - & 46  & 25 & - \\
		\midrule
		VariBAD  & \textbf{1}	 & \textbf{100} & \textbf{100} & 29 (6/20 seeds) 
											& \textbf{100} & \textbf{100} & 2 (1/20 seeds)  \\
		VariBAD  & 2				& 100 & 100 & 29 
											& 100 & 100 & 2 \\
		\midrule
		HyperX 	& \textbf{1} 	& \textbf{100} & \textbf{100} & \textbf{43} (9/20 seeds)
											& 100  & 100 & 2 (1/20 seeds) \\
		HyperX 	& 2 				& 100 & 100 & 43 
											& 100  & 100 & 2 \\
		\bottomrule
	\end{tabular}
	\vspace{0.1cm}
	\caption{
		\textbf{Meta-test success rates on the ML1 Meta-World benchmark, for the dense and the sparse reward version.} 
		$^*$Results taken from \citet{yu2019meta}. 
		$^+$Results taken from \citet{zhang2020learn}. 
		We ran VariBAD and HyperX for $5$ random seeds for dense reach/push, and $20$ seeds for dense pick-place.
		VariBAD and HyperX were trained to maximise expected online return within $2$ episodes. 
		The first (few) episodes often \emph{includes exploratory actions}, yet have higher success rate than existing methods that maximise final episodic return.
		For the sparse Pick-Place environment, in brackets we report the number of seeds that learned something.
	}
	\label{table:ml1_results}
\end{table*}

To test how our method scales up to more challenging problem settings, we evaluate it on the Meta-World benchmark \citep{yu2019meta}, where a simulated robot arm has to perform tasks.
We evaluate our method on the ML1 benchmark, of which three different versions exist: reach/push/pick-and-place (in increasing order of complexity).
In each of these, task distributions are generated by varying the starting position of the agent and the goal/object positions.

Each environment has a dense reward function that was designed such that an agent trained on a single task (i.e., fixed starting/object/goal position) can learn to solve it. 
Evaluation is done in terms of success rate (rather than return), which is a task-specific binary signal indicating whether the task was accomplished (at any moment during the rollout).
\citet{yu2019meta} proposed a sparse version of this benchmark that uses this binary success indicator, rather than the dense reward, for training. 
This sparse version was used in \cite{zhang2020learn}, on ML1-reach and ML1-push.

The agent is trained on a set of $50$ environments and evaluated on unseen environments from the same task distribution. 
In all baselines, the agent has $10$ episodes to adapt, and performance is measured in the last rollout. 
Since we consider the \emph{online adaptation} setting where we want the agent has to perform well from the start, 
we trained VariBAD and HyperX to maximise online return during the first two episodes.
This is more challenging since it includes exploratory actions.

Table \ref{table:ml1_results} shows the results for both the dense and sparse versions of ML1. 

\textbf{ML1-reach / ML1-push.}
VariBAD achieves 100\% success rate on both the dense \emph{and the sparse} version of ML1-reach and ML1-push \emph{in the first rollout}. 
Compared to other existing methods -- even MetaCURE \citep{zhang2018decoupling} which explicitly tries to deal with sparsity -- this is a significant improvement.
We confirm in our experiments that HyperX does not decrease performance and also reaches 100\% success rate on these environments. 

\textbf{ML1-pick-place.}
The environment ML1-pick-place is more challenging, because the task consists of two steps: picking up an object and placing it somewhere (where both the object and goal location differ across tasks). 
Even on the dense version, existing methods struggle. 
HyperX achieves state of the art on this task with 44.5\% success rate, suggesting HyperX can help meta-learning even when rewards are dense.
For VariBAD and HyperX we found that our agents either learn the task near perfectly (and have close to 100\% success rate in the first rollout), or not at all. 
VariBAD learned something for $6$ out of $20$ seeds, and HyperX learned something for $9$ out of $20$ seeds. 
For the sparse version of this environment, we only saw some success for $1$/$20$ seeds for both VariBAD or HyperX.
 
We suspect that the main challenge in ML1-Pick-Place is the short horizon ($150$), which does not give the agent enough time to explore during meta-training.
This is why HyperX can give some improvement even in the dense version.
In an upcoming version of Meta-World \cite{yu2019meta}, the horizon will be increased to $200$, opening up interesting opportunities for future research on sparse Pick-Place.

\subsection{Sparse 2D Navigation}\label{appendix:additional_results:point_robot}

\begin{figure*}[t]
	\centering
	\begin{subfigure}{0.33\linewidth}
		\includegraphics[width=\linewidth]{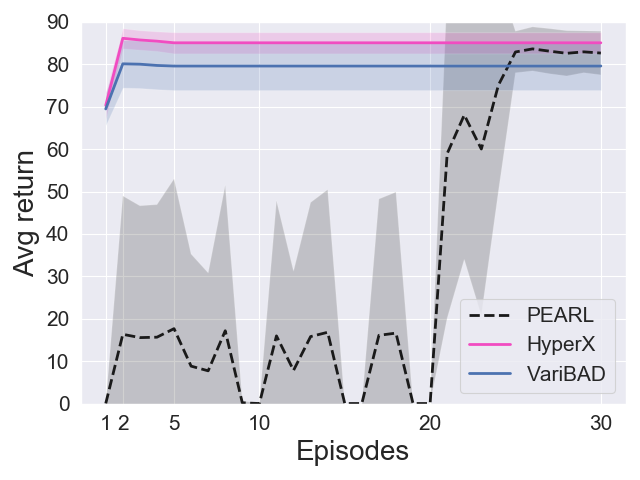}
		\caption{Meta-Test Performance}
		\label{fig:pointrobot:test_performane}
	\end{subfigure}
	\begin{subfigure}{0.33\linewidth}
		\includegraphics[width=\linewidth]{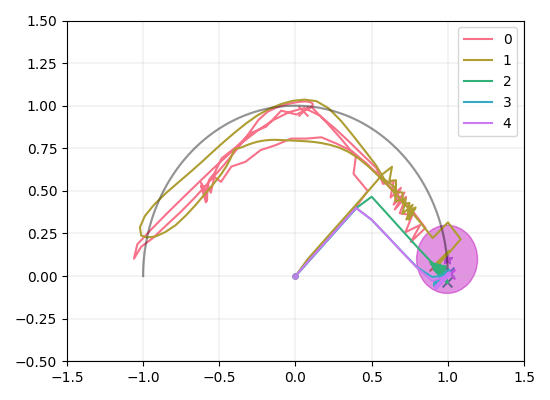}
		\caption{VariBad example rollout}
		\label{fig:behaviour_2D_navigation_varibad}
	\end{subfigure}
	\begin{subfigure}{0.33\linewidth}
		\includegraphics[width=\linewidth]{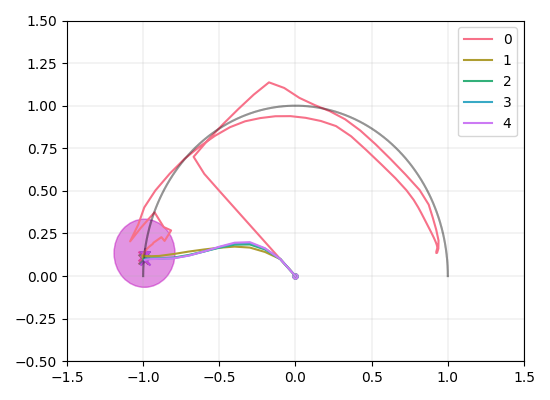}
		\caption{HyperX example rollout}
		\label{fig:behaviour_2D_navigation_hyperx}
	\end{subfigure}
	\caption{
	Meta-test performance on the Sparse 2D Navigation environment. 
	\emph{Left:} Performance averaged over the task distribution at the end of training.
	Because PEARL is not optimizing for optimal exploration, it requires many more episodes to find
	the goal.
	Both VariBad and HyperX optimise for optimal exploration and are able to quickly find the goal.
	However, VariBad's exploration is suboptimal, not covering all possible goal locations equally
	well (see middle plot), explaining the lower performance compared to HyperX.
	}
\end{figure*}

We evaluate on a Point Robot 2D navigation task used by \citet{gupta2018meta,rakelly2019efficient,humplik2019meta}.
The agent must navigate to an unknown goal sampled along the border of a semicircle of radius $1.0$, and receives a reward relative to its proximity to the goal when it is within a goal radius of $0.2$. 
Thus far, only \citet{humplik2019meta} successfully meta-learn to solve this task by meta-training with sparse rewards, though they rely on privileged information during meta-training (the goal position). 
The other methods meta-train with dense rewards and evaluate using sparse rewards.
We use a horizon of $100$ here (instead of $20$ as in the papers above) to give VariBAD and HyperX enough time to demonstrate interesting exploratory behaviour.

Figure \ref{fig:pointrobot:test_performane} shows the performance of PEARL, VariBAD, and HyperX at test time, when rolling out for $30$ episodes.
Both VariBAD and HyperX adapt to the task quickly compared to PEARL, 
but HyperX reaches slightly lower final performance.

To shed light on these performance differences, Figures \ref{fig:behaviour_2D_navigation_varibad} and \ref{fig:behaviour_2D_navigation_hyperx} visualise representative example rollouts for the meta-trained VariBAD and HyperX agents.
We picked examples where the target goals are at the end of the semi-circle, which we found are most difficult for the agents.
VariBAD (\ref{fig:behaviour_2D_navigation_varibad}) struggles to find the goal, taking several attempts to reach it. 
Once the goal is found, it does return to it but on a sub-optimal trajectory. 
By contrast, HyperX searches the space of goals more strategically, and returns to the goal faster in subsequent episodes.

\subsection{Treasure Mountain} \label{appendix:additional_results:treasure_mountain}

\begin{figure*}
	\begin{subfigure}{0.33\linewidth}
		\centering
		\includegraphics[width=\linewidth]{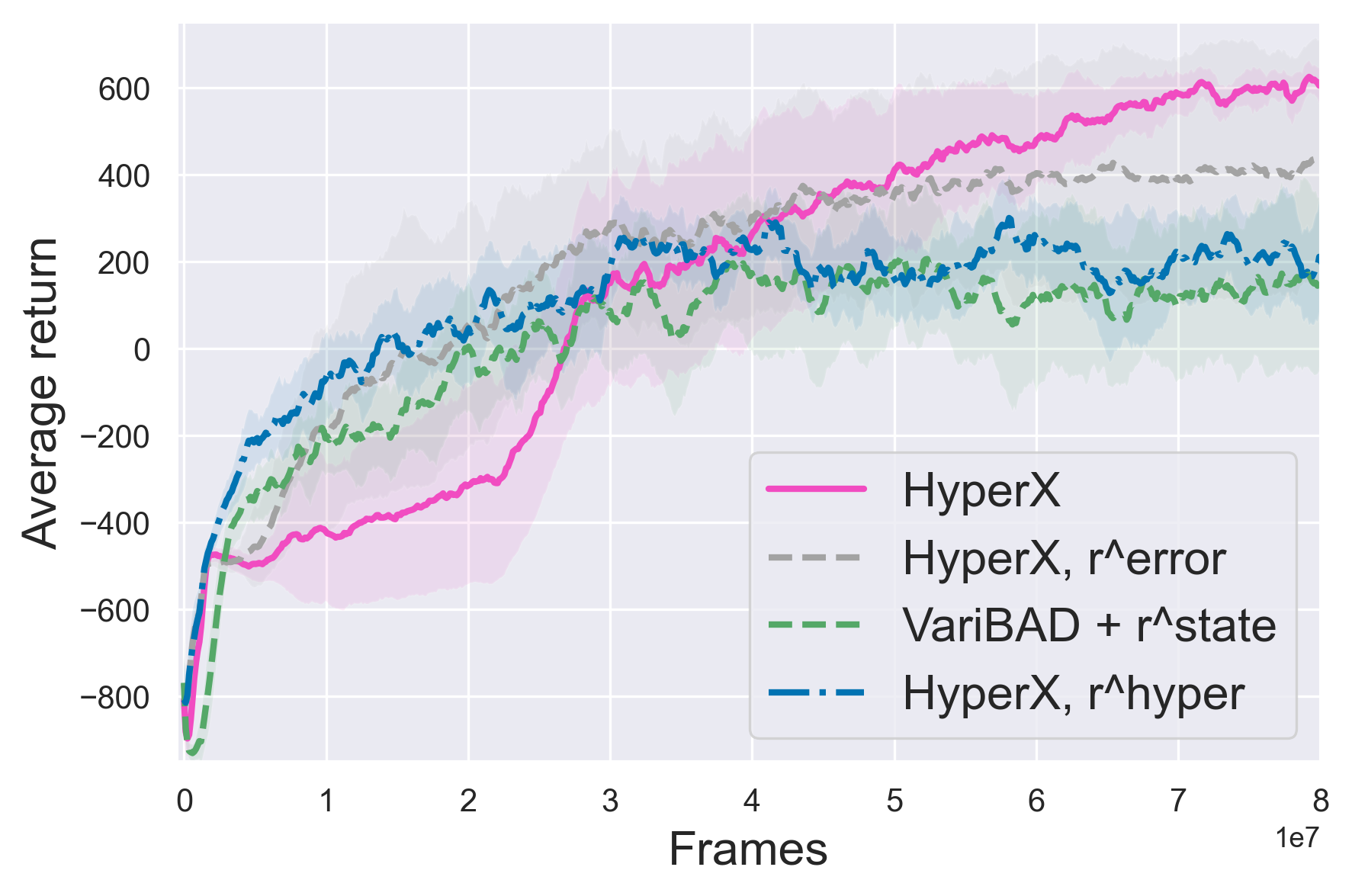}
	\caption{
		Learning curves for ablations.
	}
	\label{fig:treasure_learning_curve_ablations}
	\end{subfigure}
	\begin{subfigure}{0.33\linewidth}
	\centering
	\includegraphics[width=\linewidth]{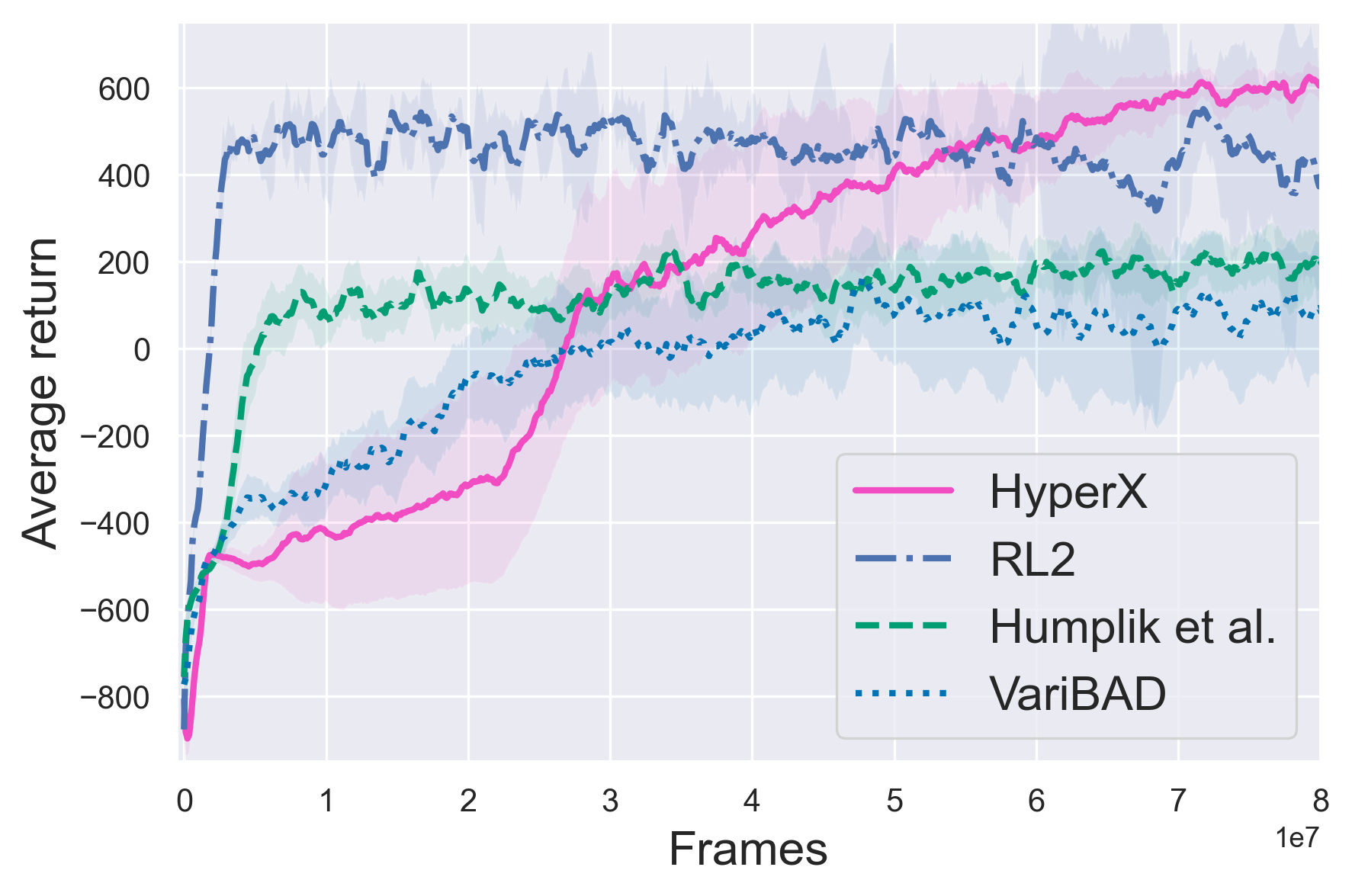}
	\caption{
		Learning curves for baselines.
	}
	\label{fig:treasure_learning_curve_baselines}
	\end{subfigure}
	\begin{subfigure}{0.15\linewidth}
		\centering
		\includegraphics[width=\linewidth]{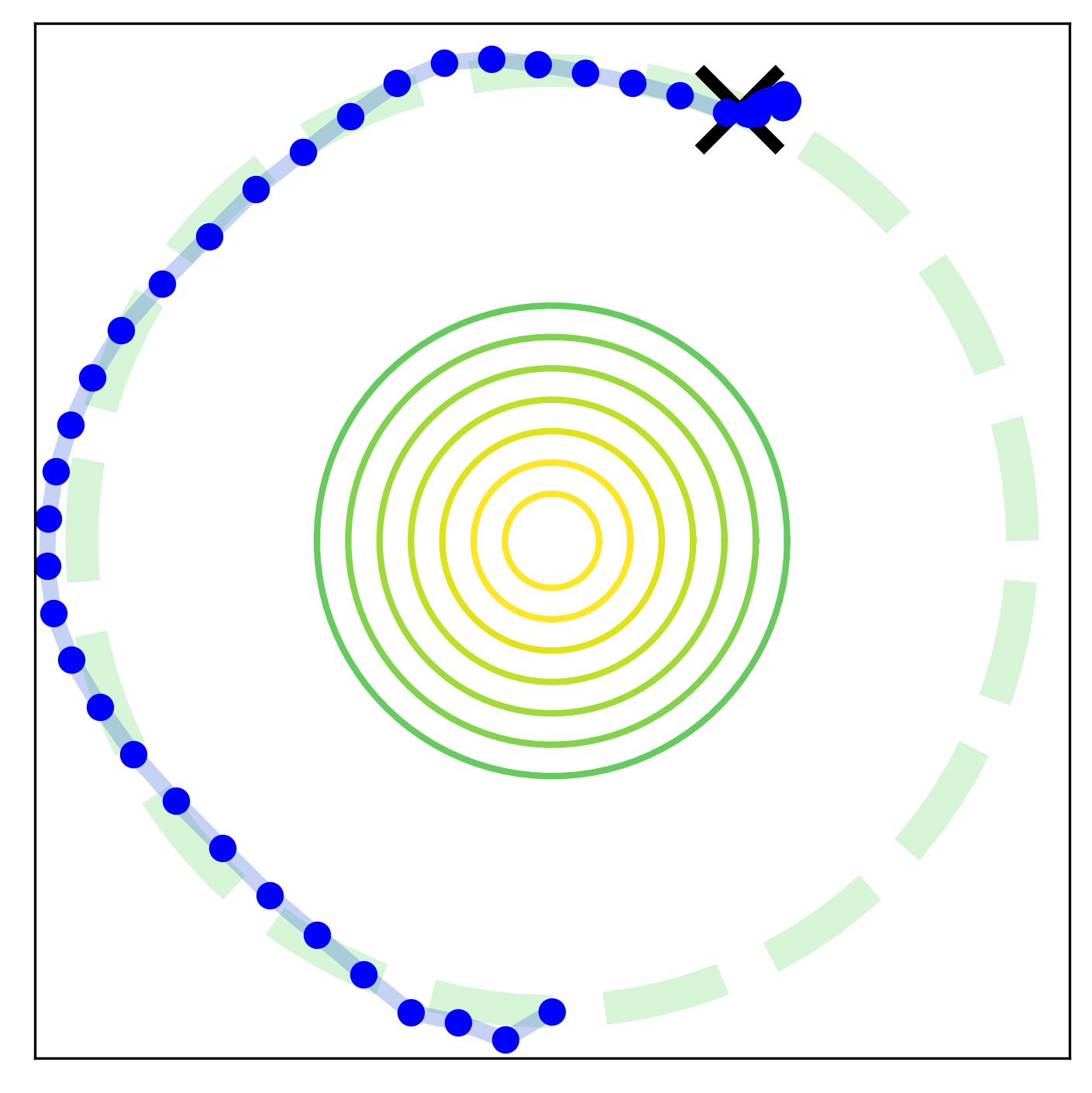}
		\caption{VariBAD}
		\label{fig:treasure_varibad_rollout}
	\end{subfigure}%
	\begin{subfigure}{0.15\linewidth}
		\centering
		\includegraphics[width=\columnwidth]{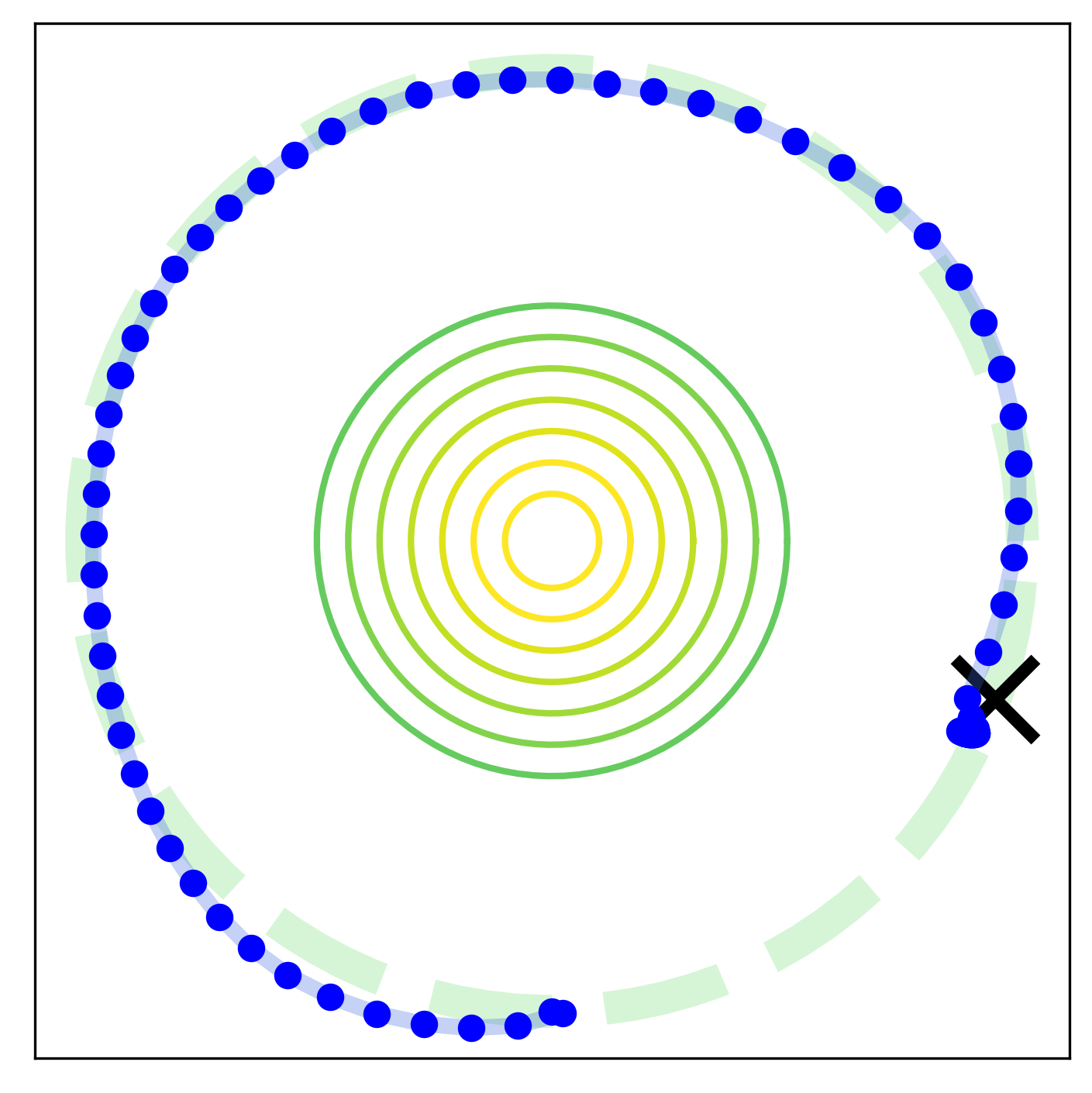}
		\caption{RL$^2$}
		\label{fig:treasure_rl2_rollout}
	\end{subfigure}
	\caption{
		\textbf{Treasure Mountain - Additional Rollouts.}
		Shown are example rollouts for the final agents of VariBAD \cite{zintgraf2020varibad} and RL$^2$ \cite{duan2016rl, wang2016learning}.
		They follow the inferior exploration strategy of walking around the circle until the treasure is found, 
		instead of climbing the mountain to directly observe the treasure and get there faster.
	}
	\label{fig:treasure_additional_rollouts}
\end{figure*}

\textbf{Ablations.}
Figure \ref{fig:treasure_learning_curve_ablations} shows the learning curves for the HyperX, in comparison to ablating different exploration bonuses.
When using only the hyper-state novelty bonus $r^{hyper}$, HyperX learns the inferior strategy of walking in a circle: it has no incentive to go up the mountain early in training (because beliefs there are meaningless because the VAE has not learned yet to interpret the hint) and stars avoiding the mountain. 
When using only the VAE reconstruction error bonus $r^{error}$, the agent learns the superior strategy of walking up the mountain to see the goal location 70\% of the time (7/10 seeds). 
In contrast, HyperX, which uses both exploration bonuses, learns the superior strategy for all $10$ seeds.
Lastly, we tested VariBAD with a simple state novelty exploration bonus: this again learns the inferior circle-walking strategy only, because it quickly learns to avoid the mountain top.

\textbf{Baselines - Performance.}
Figure \ref{fig:treasure_learning_curve_baselines} shows the learning curves for HyperX and VariBAD (discussed in Sec \ref{sec:experiments:treasure}), as well as additional baselines $RL^2$ \citep{duan2016rl, wang2016learning} (which is a model-free method where the policy is a recurrent network that gets previous actions and rewards as inputs in addition to the environment state) and the Belief Learning method of \citet{humplik2019meta} (which uses privileged information -- the goal position -- during meta-training).
Both these baselines also only learn the inferior circle-walking strategy, because the correct incentives for meta-exploration are missing.

\textbf{Baselines - Behaviour.}
Figures \ref{fig:treasure_varibad_rollout} and \ref{fig:treasure_rl2_rollout} show meta-test time behaviour of VariBAD and RL$^2$: both methods learn to walk in a circle until the goal is found.
This was consistent across all ($10$) seeds.

\subsection{Sparse CheetahDir} \label{appendix:cheetah_results}

\begin{figure*}
		\begin{subfigure}{0.33\linewidth}
			\centering
			\includegraphics[width=\linewidth]{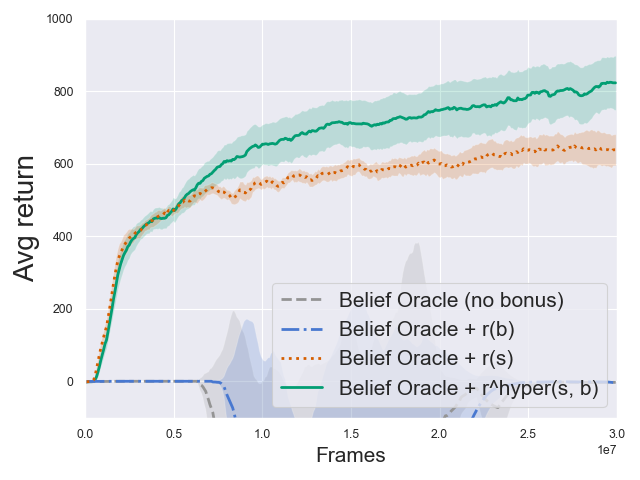}
			\caption{Belief Oracle.}
			\label{subfig:belief_oracle_learning_curve}
		\end{subfigure}
	\begin{subfigure}{0.33\linewidth}
		\centering
		\includegraphics[width=\linewidth]{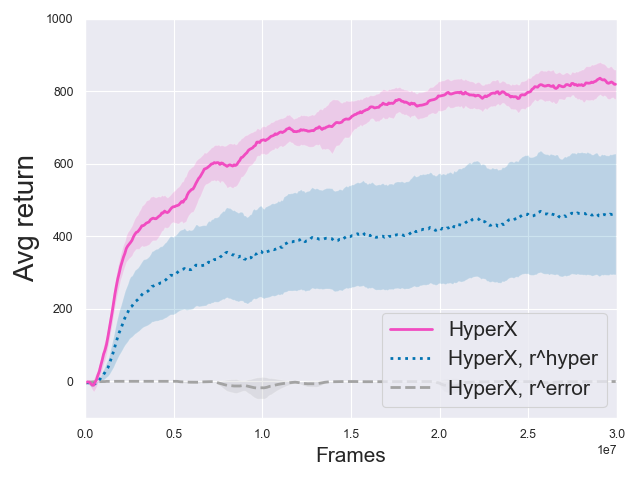}
		\caption{HyperX}
		\label{subfig:variBAD_learning_curve}
	\end{subfigure}
	\begin{subfigure}{0.33\linewidth}
	\centering
	\includegraphics[width=\linewidth]{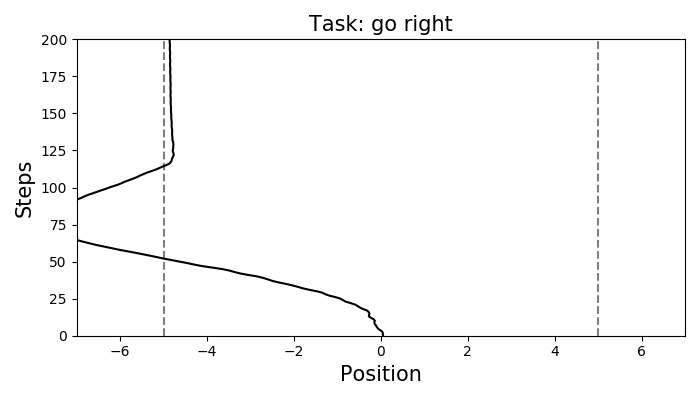}
	\caption{Behaviour of a policy which failed to learn Bayes-optimal behaviour. We observe such behaviour often when training HyperX with the reward bonus on the hyper-states only, $r^{hyper}(b, s)$.}
	\label{fig:behaviour_varibad}
	\end{subfigure}
	\caption{
		\textbf{HalfCheetahDir: Additional Results.}
		Learning curves for the Belief Oracle (A) and HyperX (B), with and without reward bonus, averaged over $20$ seeds..
	}
	\label{fig:cheetah:learning_curves}
	\end{figure*}

Figure \ref{fig:cheetah:learning_curves} shows the learning curves for the Sparse CheetahDir experiments, with $95\%$ confidence intervals (over $20$ seeds). 
Fig \ref{subfig:belief_oracle_learning_curve} shows this for the Belief Oracle, with different exploration bonuses. 
Fig \ref{subfig:variBAD_learning_curve} shows this for HyperX, with different exploration bonuses.

Figure \ref{fig:behaviour_varibad} shows example behaviour of a suboptimal policy at test time.
The agent returns back into the zero-reward zone after realising that the task was not "go left", but stays in there instead of behaving optimally, which is going further to the right and into the dense reward area beyond the sparse interval border.

\subsection{Sparse MuJoCo AntGoal} \label{appendix:antgoal_results}

\begin{figure*}
	\begin{subfigure}{0.33\linewidth}
		\centering		
		\includegraphics[width=\linewidth]{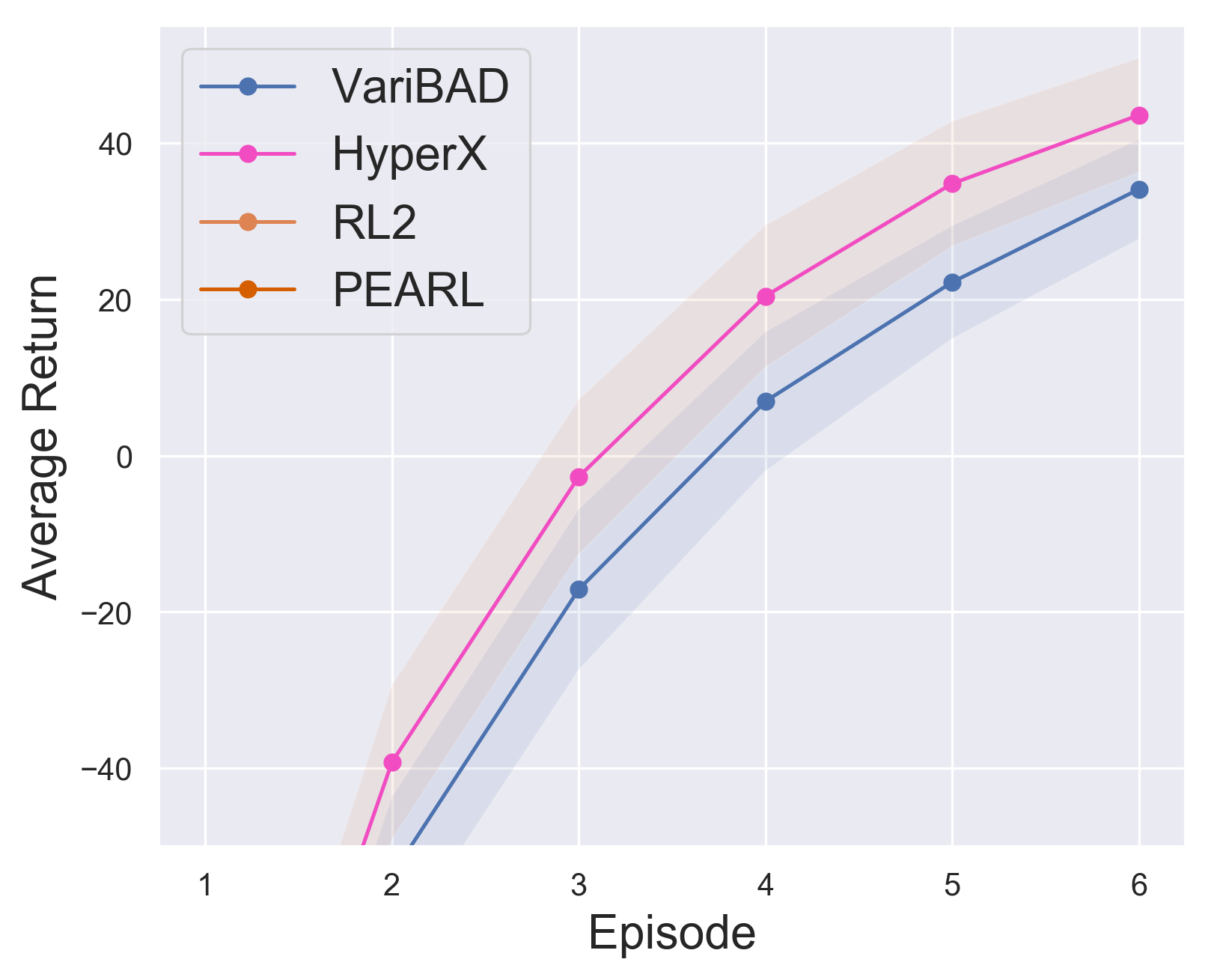}
		\caption{
			Return per episode at meta-test time (standard error shaded). RL2 and PEARL do not learn to solve the task and achieve a reward of around -150 per episode.
		}
		\label{fig:antgoal:ep_return}
	\end{subfigure}
	\begin{subfigure}{0.33\linewidth}
		\centering
		\includegraphics[width=\linewidth]{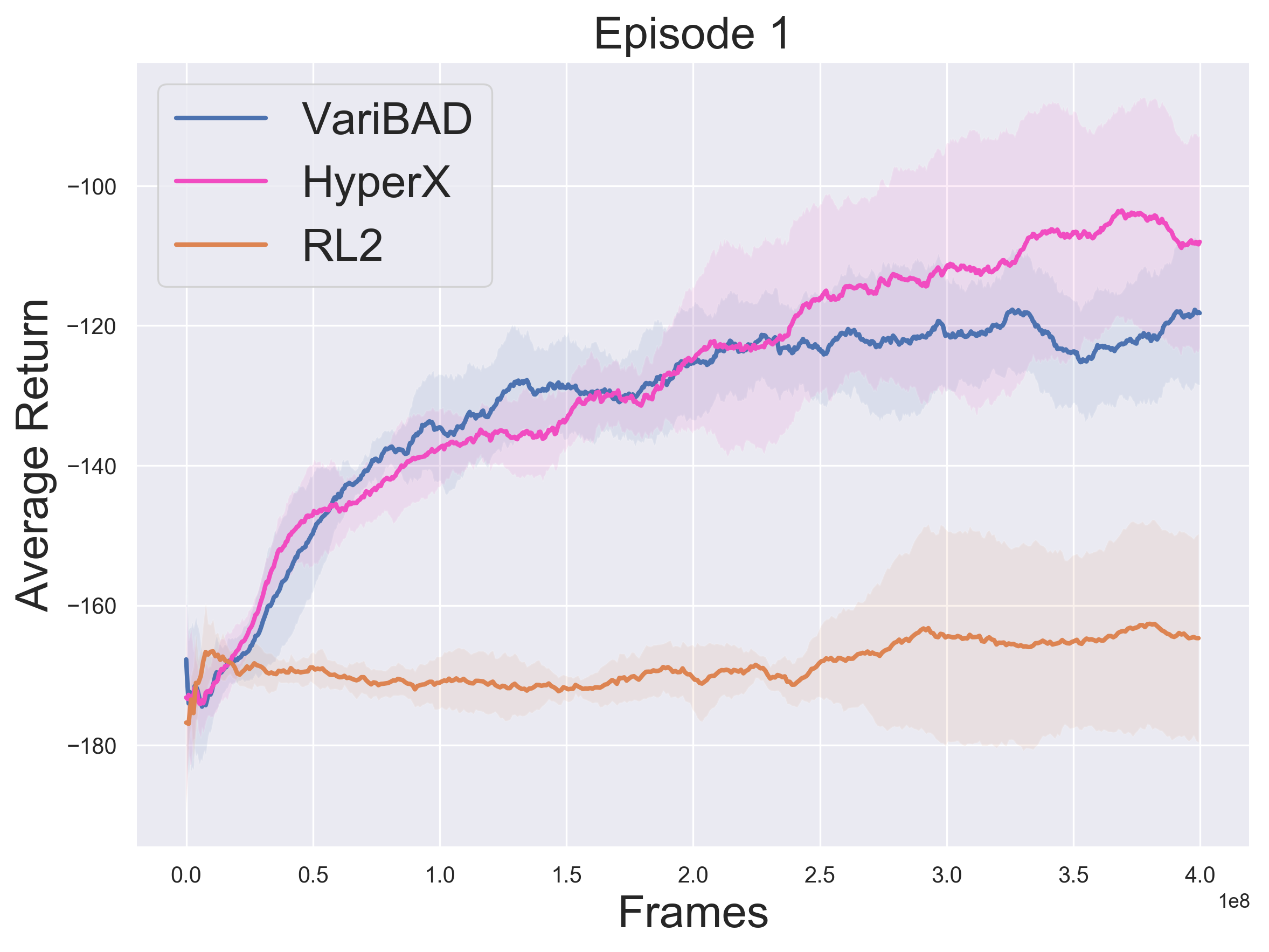}
		\caption{
			Learning curve (return in ep 1)
		}
		\label{subfig:antgoal:learning_curve_ep1}
	\end{subfigure}
	\begin{subfigure}{0.33\linewidth}
		\centering
		\includegraphics[width=\linewidth]{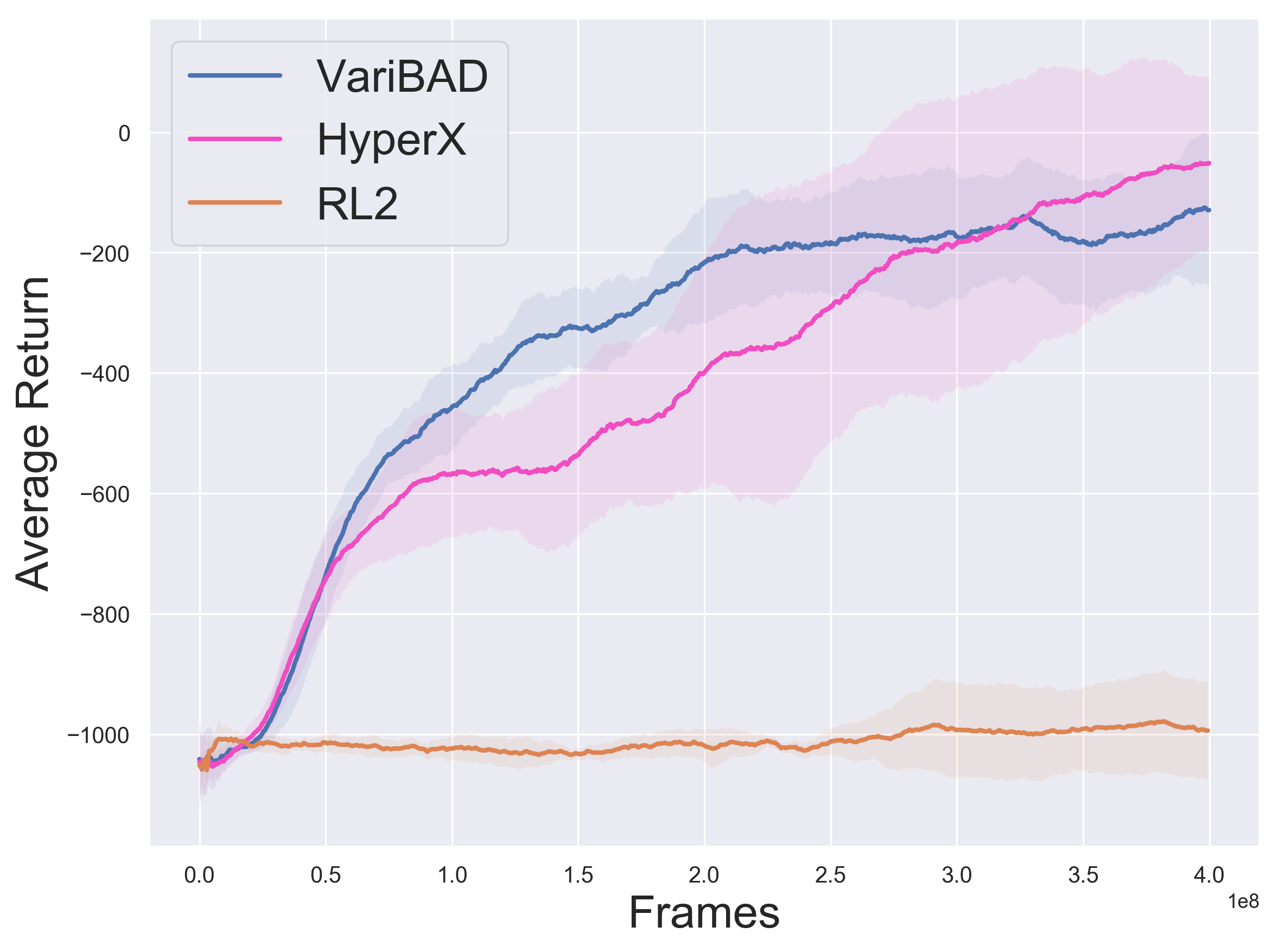}
		\caption{
			Learning Curve (sum of returns in ep 1-6)
		}
		\label{subfig:antgoal:learning_curve_combined}
	\end{subfigure}
	\caption{\textbf{Sparse AntGoal: Additional Plots.} ($10$ seeds).}
	\label{subfig:antgoal:additional_results}
\end{figure*}

In addition to the main results in the paper (Sec \ref{sec:experiments:ant_goal}) we provide additional experimental results here.

\begin{figure}
	\centering
	\begin{subfigure}{0.49\columnwidth} 
		\includegraphics[width=\columnwidth]{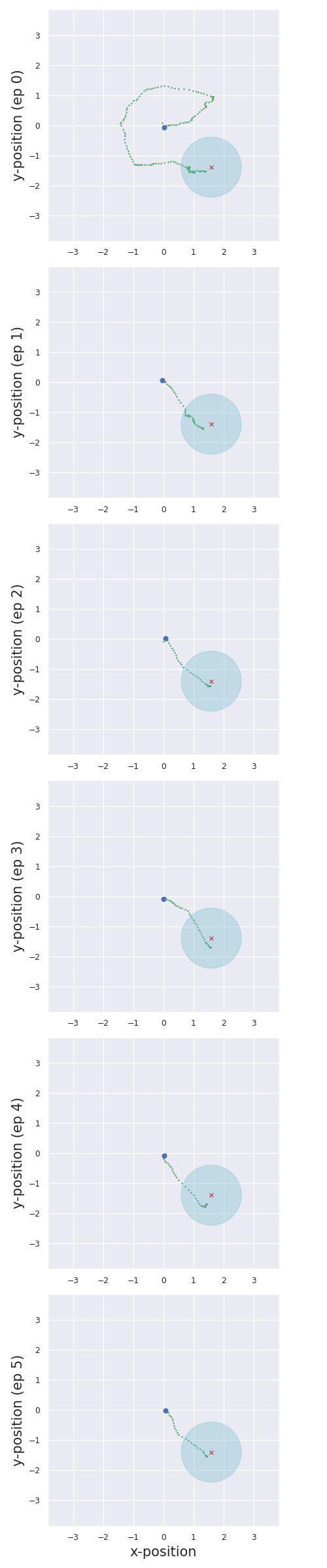}
	\end{subfigure}
	\hfill
	\begin{subfigure}{0.49\columnwidth}
		\includegraphics[width=\columnwidth]{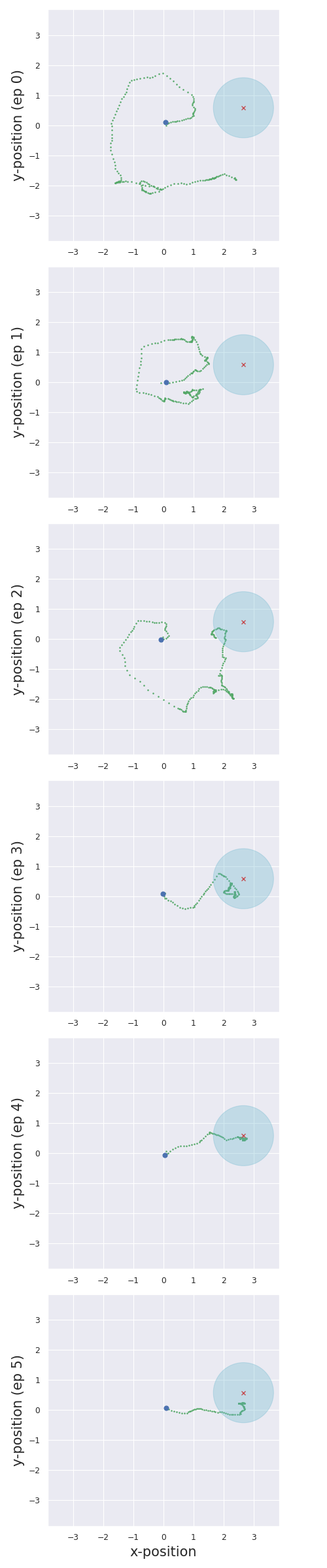}
	\end{subfigure}
	\caption{HyperX Example Rollouts}
	\label{fig:antgoal:hyperx_rollouts}
\end{figure}

\begin{figure}
	\centering
	\begin{subfigure}{0.49\columnwidth}
		\includegraphics[width=\columnwidth]{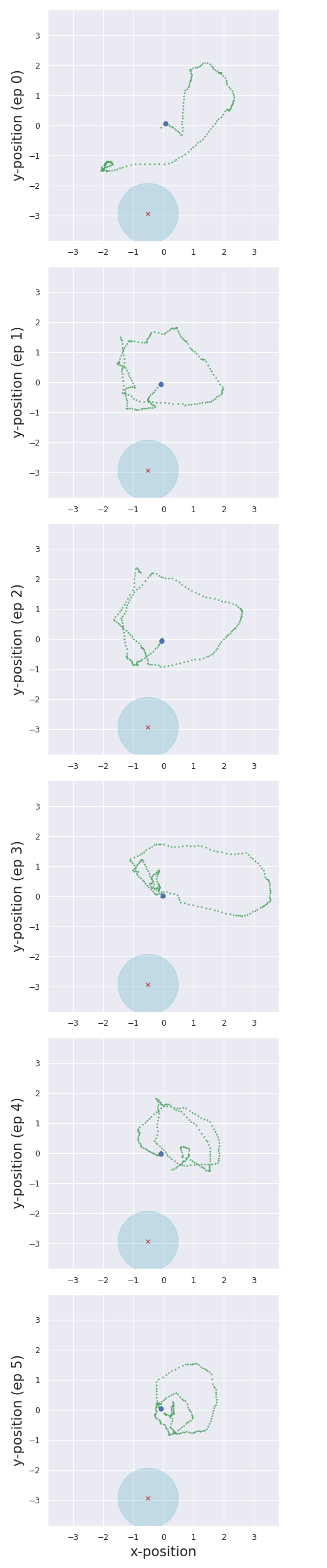}
	\end{subfigure}
	\hfill
	\begin{subfigure}{0.49\columnwidth}
		\includegraphics[width=\columnwidth]{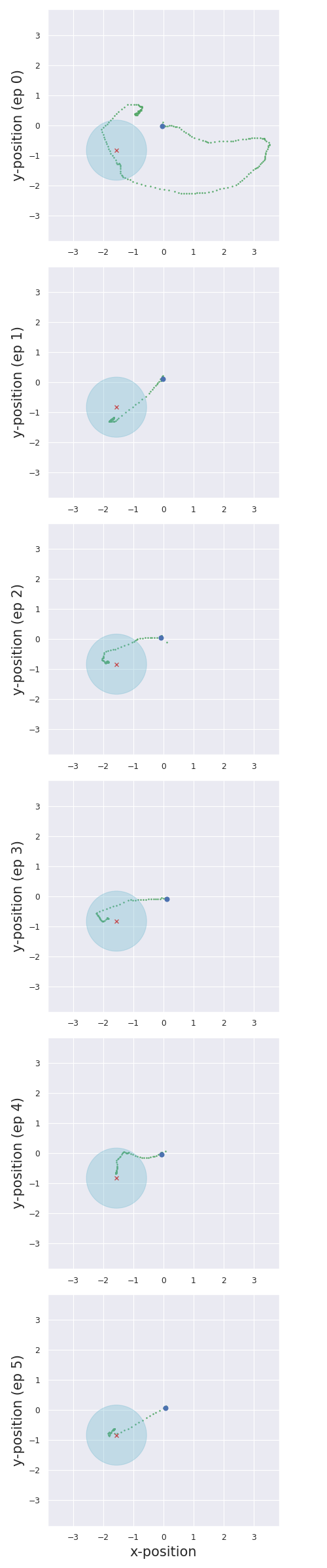}
	\end{subfigure}
	\caption{VariBAD Example Rollouts}
	\label{fig:antgoal:varibad_rollouts}
\end{figure}

Figure \ref{fig:antgoal:ep_return} shows the returns achieved by the agents across different episodes. 
Figures \ref{subfig:antgoal:learning_curve_ep1} show the learning curves for the returns during the \emph{first} episode, with $95\%$ confidence intervals (shaded areas, $10$ seeds). 
Figure \ref{subfig:antgoal:learning_curve_combined} shows the combined learning curves, comprising of all 6 episodes, with $95\%$ confidence intervals (shaded areas, $10$ seeds).
Figures \ref{fig:antgoal:varibad_rollouts} and \ref{fig:antgoal:hyperx_rollouts} show example rollouts for VariBAD and HyperX.

\textbf{Dense AntGoal.}
We also evaluated HyperX on the dense AntGoal environment. 
VariBAD and HyperX were trained to maximise performance within a single episode. 
PEARL was trained with the default hyperparameters provided by the open-sourced code of the authors.
The results are:: VariBAD: -123 (Episode 1), HyperX: -127 (Episode 1), PEARL: -200 (Episode 6).
This confirms that HyperX does not impact performance, but that there is also not much room for improvement.

\newpage 
\section{Implementation Details} \label{appendix:implementation_details}

The source code is available as additional supplementary material.
In this section, we provide the 
environment specifications (\ref{appendix:implementation_details:environment_specs}), 
runtimes (\ref{appendix:implementation_details:runtimes}), 
and hyperparameters (\ref{appendix:implementation_details:hyperparameters}).

\subsection{Environment Specifications} \label{appendix:implementation_details:environment_specs}

In this section we provide additional details on the environments that were used in the main paper. 
Implementation of these environments are in the provided source code. 

\subsubsection{Treasure Mountain} \label{appendix:environments:mountain}

This environment is implemented as follows.
The treasure can be anywhere along a circle of radius $1$.
Within that circle is a mountain -- implemented as another circle with radius $0.5$.
The horizon is $100$ and there are no resets.
The agent always starts at he bottom of the circle.
It receives a reward of $10$ when it is within a Euclidean distance of $0.1$ within the treasure (the treasure does not disappear, so it keeps receiving this reward if it stays there).
It receives a penalty for climbing the mountain, given my $-5.5 + ||(x,y)||_2$ where $(x,y)$ is the agent's position (the mountain center is $0,0$, and the mountain radius $0.5$).
If not at the treasure or on the mountain, the agent gets a timestep penalty of at least $-5$, which increases as the agent walks further outside the outer circle (to discourage it from walking too far).
The agent cannot walk outside $[-1.5, 1.5]$ in either direction.

The observations of the agent are $4$D and continuous. The first two dimensions are the agent's $(x,y)$-position. 
The last two dimensions are zero if the agent is not on the mountain top, and are the $(x,y)$-coordinates of the treasure when the agent is on the mountain top (within a radius of $0.1$).
The agent's actions are the (continuous) stepsize it takes in $(x,y)$-direction, bounded in $[-0.1, 0.1]$.

\subsubsection{Multi-Stage Gridworld}

The layout of this environment is depicted in Fig \ref{fig:room}.
It consist of three rooms which are of size $3\times3$ grid, and corridors that connect the rooms of length $3$.
The environment state is the $(x,y)$ position of the agent, unnormalised. 
There are five available actions: \emph{no-op, up, right, down, left}.

Three (initially unknown) goals (G1-G3) are placed in corners of rooms: G1 in the middle room, G2 in the room that is on the side where G1 was placed, and G3 in the middle room (but not where G1 was placed).
The agent always starts in the middle of the centre room and has $H=50$ steps.
The goals provide increasing rewards, i.e. $r_{1}=1$, $r_{2}=10$ and $r_{3}=100$, but are only sequentially unlocked; G2 ($r_{2}$) is only available after G1 has been reached; G3 ($r_{3}$) is only available after G2 has been reached.
The environment is partially observable \citep{poupart2008model, cai2009learning} as the agent only observes its position in the environment and not which goals are unlocked.
If the agent is not on an (available) goal it gets $r=-0.1$. 
When the agent stands on a goal, it keeps receiving the respective reward while standing there (the goal does \emph{not} disappear).
The best strategy is to search the first room for G1, then search the appropriate room for G2, and then return to the middle room to find G3.

\subsection{Sparse HalfCheetahDir}

We use the commonly used HalfCheetahDir meta-learning benchmark (based on code of \citet{zintgraf2020varibad}), and sparsify it as follows.
If the agent's x-position is within $[-5, 5]$ it only gets the control penalty; otherwise it gets the standard dense reward comprised of the sum of the control penalty and the $1$D velocity in the correct direction. 

\subsection{Sparse MuJoCo AntGoal}

We use the commonly used AntGoal meta-learning benchmark (based on code of \citet{rakelly2019efficient}), and sparsify it as follows. 
We extend the environment's state space by including the x and y-position of the agent's torso. 
In the original AntGoal, the goal is sampled from within a circle of radius of $3$ with a higher chance of the goal being sampled away from the centre of the circle. 
Unlike the dense version where the agent receives a dense goal-related reward at all times, our sparse AntGoal only receives goal-related rewards when within a radius of $1$ of the goal.

The agent receives at all time a control penalty and contact forces penalty. When outside the goal circle, the agent receives an additional constant negative reward that is equivalent to the negative goal radius, i.e. $-1$. When within the goal circle, the agent receives a reward of $1$ for being within the goal circle and a penalty equivalent to the negative distance to the goal, essentially encouraging the agent to walk towards the centre of the goal circle.

\subsection{Meta-World}

We use the official version of Meta-World as provided by \citet{yu2019meta} at \url{https://github.com/rlworkgroup/metaworld}.
As suggested by \citet{yu2019meta} and as tested in \citet{zhang2020learn}, for the sparse version of this environment, we use the \emph{success} criterion which the environment returns, and give the agent a reward of $0$ if \emph{success=False} and a reward of $1$ if \emph{success=True}.
The success criterion depends on the environment; in `Reach' for example it is \emph{true} if the agent put its gripper close to the (initially unknown) goal position, and \emph{false} otherwise.
For evaluation, we report `Success' if the agent was successful at any moment during an episode, following the evaluation protocol proposed by \citet{yu2019meta}.

\subsection{Runtimes} \label{appendix:implementation_details:runtimes}

Table \ref{table:runtimes} shows the runtimes for our experiments. 
Unless otherwise stated, we used a NVIDIA GeForce GTX 1080 GPU. 
These runtimes should serve as a rough estimate, and can vary depending on hardware and concurrent processes.

\begin{table}[h]
	\centering
	\begin{tabular}{l|c|c}
		\toprule
		Environment  					& Frames 		& Runtime (ca.) \\
		\midrule
		Treasure Mountain 			& $8e{+}7$ 		& $35$h \\
		Multi-Stage Gridworld 		& $1e{+}8$ 		& $65$h (CPU) \\
		Sparse HalfCheetahDir 		& $3e{+}7$	& $20$h (CPU) \\
		Sparse AntGoal 					& $4e{+}8$ 	& $65$h \\
		Meta-World 						& $5e{+}7$ 		& $45$h \\
		Sparse $2$D Navigation 	& $5e{+}7$ & $12$h \\
		\bottomrule
	\end{tabular}
	\caption{}
	\label{table:runtimes}
\end{table}

\subsection{Hyperparameters} \label{appendix:implementation_details:hyperparameters}

We train the policy using PPO, and we add the intrinsic bonus rewards to the extrinsic environment reward and use the sum when learning with PPO.
We normalise the intrinsic and extrinsic rewards separately by dividing by a rolling estimate of the standard deviation. 

On the next two pages we show the hyperparameters used for the policy, the VAE, and the exploration bonuses.
Hyperparameters were selected using a simple (non-exhaustive) gridsearch.

For the MuJoCo environments, we only used the relevant state information for the RND hyper-state bonus (the $x$-axis for HalfCheetahDir, and the $x$-$y$-position for AntGoal).

\paragraph{RND Hyperparameter Sensitivity.}
To assess how sensitive HyperX to choices of hyperparameters that affect the hyperstate exploration bonus, we evaluated it on a range of different choices, shown in Table \ref{tab:additional_results}.
There is little sensitivity to architecture depth and batchsize, as well as to the output dimension of the RND networks.
Performance is stable for learning rates $10^{-3}{-}10^{-6}$ (possibly because we use the Adam optimiser), but we found that the best frequency ($freq$) at which the RND network is updated to be environment dependent.
Performance is sensitive to the scaling factor ($wsi$ in the table) for the initial prior network weights. We used a scaling factor of $10$ in our experiments, and found that too small or too large scaling factors can hurt performance.
An interesting direction for future work is to find more principled ways to guide the choice of the hyperparameters that are particularly sensitive to the exploration and across environments. 

\begin{table}[h]
	\centering
	\begin{tabular}{l|c}
		\toprule
		RND $dim_{out}=32$ (default $128$) & 737 \\
		RND $dim_{out}=256$ (default $128$) & 812 \\
		\hline
		RND $depth=1$ (default $2$) & 794 \\
		RND $depth=3$ (default $2$) & 814 \\
		\hline
		RND $batchsize=32$ (default $128$) & 856 \\
		RND $batchsize=256$ (default $128$) & 867 \\
		\hline
		RND $lr=1e-2$ (default $1e-4$) & 108 \\
		RND $lr=1e-3$ (default $1e-4$) & 883 \\
		RND $lr=1e-5$ (default $1e-4$) & 845 \\
		RND $lr=1e-6$ (default $1e-4$) & 766 \\
		\hline
		RND $wsi=1$ (default $10$) & 597 \\
		RND $wsi=5$ (default $10$) & 766 \\
		RND $wsi=15$ (default $10$) & 533 \\
		\bottomrule
	\end{tabular}
	\caption{Additional Sparse CheetahDir Results, for different RND hyperparameter settings (averaged over three seeds). $wsi$ stands for weight scale initialisation of the fixed random prior network.}
	\label{tab:additional_results}
\end{table}

\newpage
\ \\ 
\newpage 

\rotatebox{90}{
\begin{tabular}{lllllllll}
	\toprule
	{} &    Treasure & GridWorld &  CheetahDir &          AntGoal &       PointRobot &   ML1-Reach &    ML1-Push & Ml1-Pick-Place \\
	\midrule
	max\_rollouts\_per\_task        &           1 &         1 &           1 &                6 &                3 &           2 &           2 &              2 \\
	policy\_state\_embedding\_dim   &        None &        32 &          32 &               64 &               64 &          64 &          64 &             64 \\
	policy\_latent\_embedding\_dim  &        None &        32 &          32 &               64 &               64 &          64 &          64 &             64 \\
	norm\_state\_for\_policy        &       False &      True &        True &             True &             True &        True &        True &           True \\
	norm\_latent\_for\_policy       &       False &      True &       False &             True &             True &        True &        True &           True \\
	norm\_rew\_for\_policy          &        True &      True &        True &            False &             True &        True &        True &           True \\
	norm\_actions\_pre\_sampling    &       False &     False &       False &             True &             True &       False &       False &          False \\
	norm\_actions\_post\_sampling   &        True &     False &       False &            False &            False &       False &       False &           True \\
	norm\_rew\_clip\_param          &         100 &       NaN &         NaN &           100000 &           100000 &      100000 &      100000 &         100000 \\
	policy\_layers                &  [128, 128] &      [64] &  [128, 128] &  [128, 128, 128] &  [128, 128, 128] &  [128, 128] &  [128, 128] &     [128, 128] \\
	policy\_activation\_function   &        tanh &      tanh &        tanh &             tanh &             tanh &        tanh &        tanh &           tanh \\
	policy\_initialisation        &  orthogonal &     normc &       normc &            normc &            normc &       normc &       normc &          normc \\
	policy\_anneal\_lr             &       False &     False &       False &            False &            False &       False &       False &          False \\
	policy                       &         ppo &       ppo &         ppo &              ppo &              ppo &         ppo &         ppo &            ppo \\
	policy\_optimiser             &        adam &      adam &        adam &             adam &             adam &        adam &        adam &           adam \\
	ppo\_num\_epochs               &           2 &         8 &           2 &                2 &                2 &           2 &           2 &              2 \\
	ppo\_num\_minibatch            &           8 &         4 &           4 &                8 &                8 &           8 &           8 &              8 \\
	ppo\_clip\_param               &        0.05 &      0.05 &         0.1 &              0.1 &              0.1 &         0.1 &         0.1 &            0.1 \\
	lr\_policy                    &      0.0007 &    0.0007 &      0.0007 &           0.0003 &           0.0007 &      0.0007 &      0.0007 &         0.0007 \\
	num\_processes                &          16 &        16 &          16 &               16 &               16 &           8 &           8 &              8 \\
	policy\_num\_steps             &         150 &        50 &         200 &             1200 &              600 &         600 &         600 &            600 \\
	policy\_eps                   &       1e-08 &     1e-05 &       1e-08 &            1e-08 &            1e-08 &       1e-08 &       1e-08 &          1e-08 \\
	policy\_value\_loss\_coef       &         0.5 &       0.5 &         0.5 &              0.5 &              0.5 &         0.5 &         0.5 &            0.5 \\
	policy\_entropy\_coef          &       0.001 &       0.1 &      0.0001 &            0.001 &            0.001 &       0.001 &       0.001 &          0.001 \\
	policy\_gamma                 &        0.97 &      0.98 &        0.97 &             0.99 &             0.99 &        0.97 &        0.97 &           0.97 \\
	policy\_use\_gae               &        True &      True &        True &             True &             True &        True &        True &           True \\
	policy\_tau                   &         0.9 &      0.95 &         0.9 &              0.9 &              0.9 &         0.9 &         0.9 &            0.9 \\
	use\_proper\_time\_limits       &        True &     False &        True &             True &             True &        True &        True &           True \\
	vae\_squash\_targets           &        True &       NaN &         NaN &             True &             True &         NaN &         NaN &            NaN \\
	lr\_vae                       &       0.001 &     0.001 &       0.001 &            0.001 &            0.001 &       0.001 &       0.001 &          0.001 \\
	size\_vae\_buffer              &       10000 &    100000 &       10000 &            10000 &            10000 &       10000 &       10000 &          10000 \\
	precollect\_len               &         100 &      5000 &         500 &            50000 &             5000 &        5000 &        5000 &           5000 \\
	vae\_batch\_num\_trajs          &          15 &        25 &          10 &               10 &               10 &          10 &          10 &             10 \\
	tbptt\_stepsize               &        None &      None &        None &               50 &               50 &        None &        None &           None \\
	vae\_subsample\_elbos          &        None &      None &        None &               50 &               50 &          50 &          50 &             50 \\
	vae\_subsample\_decodes        &        None &      None &        None &             None &               50 &          50 &          50 &             50 \\
	vae\_avg\_elbo\_terms           &        True &     False &        True &             True &            False &       False &       False &          False \\
	vae\_avg\_reconstruction\_terms &       False &     False &       False &            False &            False &       False &       False &          False \\
	\bottomrule
\end{tabular}
}

\newpage 
\ \\ 
\newpage

\rotatebox{90}{
\begin{tabular}{lllllllll}
	\toprule
	{} &    Treasure &   GridWorld &  CheetahDir &     AntGoal &  PointRobot &      ML1-Reach &    ML1-Push & Ml1-Pick-Place \\
	\midrule
	num\_vae\_updates                 &           1 &           1 &           1 &          10 &           3 &              1 &           1 &              3 \\
	pretrain\_len                    &           0 &           0 &           0 &           0 &           0 &              0 &           0 &              0 \\
	kl\_weight                       &         1.0 &         0.1 &         1.0 &         1.0 &         1.0 &            1.0 &         1.0 &            1.0 \\
	action\_embedding\_size           &          16 &           0 &          16 &          16 &          16 &             16 &          16 &             16 \\
	state\_embedding\_size            &          32 &          32 &          32 &          32 &          32 &             32 &          32 &             32 \\
	reward\_embedding\_size           &          16 &           8 &          16 &          16 &          16 &             16 &          16 &             16 \\
	encoder\_layers\_before\_gru       &          [] &          [] &          [] &          [] &          [] &             [] &          [] &             [] \\
	encoder\_gru\_hidden\_size         &         128 &         128 &         128 &         128 &         128 &            128 &         128 &            128 \\
	encoder\_layers\_after\_gru        &          [] &          [] &          [] &          [] &          [] &             [] &          [] &             [] \\
	latent\_dim                      &          25 &          10 &           5 &           5 &           5 &              5 &           5 &              5 \\
	decode\_reward                   &        True &        True &        True &        True &        True &           True &        True &           True \\
	normalise\_rew\_targets           &        True &         NaN &         NaN &       False &       False &           True &        True &           True \\
	rew\_loss\_coeff                  &         1.0 &         1.0 &         1.0 &         1.0 &         1.0 &            1.0 &         1.0 &            1.0 \\
	input\_prev\_state                &        True &       False &        True &        True &        True &           True &        True &           True \\
	input\_action                    &        True &       False &        True &        True &        True &           True &        True &           True \\
	reward\_decoder\_layers           &    [64, 32] &    [64, 64] &    [64, 32] &    [64, 32] &    [64, 32] &       [64, 32] &    [64, 32] &  [128, 64, 32] \\
	decode\_state                    &        True &       False &       False &       False &       False &          False &       False &           True \\
	state\_loss\_coeff                &         1.0 &         1.0 &         1.0 &         1.0 &         1.0 &            1.0 &         1.0 &            1.0 \\
	state\_decoder\_layers            &    [64, 32] &    [32, 32] &    [64, 32] &    [64, 32] &    [64, 32] &  [128, 64, 32] &    [64, 32] &  [128, 64, 32] \\
	rlloss\_through\_encoder          &       False &       False &       False &       False &       False &          False &       False &          False \\
	intrinsic\_rew\_normalise\_rewards &        True &        True &        True &        True &        True &           True &        True &           True \\
	intrinsic\_rew\_clip\_rewards      &        None &        10.0 &        None &        10.0 &        None &           10.0 &        10.0 &           10.0 \\
	rpf\_weight\_hyperstate           &         1.0 &        10.0 &         1.0 &         5.0 &         0.1 &            1.0 &         1.0 &            1.0 \\
	intrinsic\_rew\_anneal\_weight     &        True &        True &        True &        True &        True &           True &        True &           True \\
	intrinsic\_rew\_for\_vae\_loss      &        True &        True &        True &        True &        True &           True &        True &           True \\
	intrinsic\_weight\_vae\_loss       &         1.0 &         1.0 &         1.0 &         1.0 &         1.0 &            1.0 &         1.0 &            1.0 \\
	lr\_rpf                          &      0.0001 &      0.0001 &      0.0001 &      0.0001 &      0.0001 &         0.0001 &      0.0001 &         0.0001 \\
	rpf\_batch\_size                  &         128 &         128 &         128 &         128 &         128 &            128 &         128 &            128 \\
	rpf\_update\_frequency            &           1 &           1 &           1 &           3 &           1 &              1 &           1 &             50 \\
	size\_rpf\_buffer                 &       10000 &    10000000 &       10000 &    10000000 &       10000 &          10000 &       10000 &          10000 \\
	rpf\_output\_dim                  &         128 &         128 &         128 &         128 &         128 &            128 &         128 &            128 \\
	layers\_rpf\_prior                &  [256, 256] &  [256, 256] &  [256, 256] &  [256, 256] &  [256, 256] &     [256, 256] &  [256, 256] &     [256, 256] \\
	layers\_rpf\_predictor            &  [256, 256] &  [256, 256] &  [256, 256] &  [256, 256] &  [256, 256] &     [256, 256] &  [256, 256] &     [256, 256] \\
	rpf\_use\_orthogonal\_init         &       False &       False &       False &       False &       False &          False &       False &          False \\
	rpf\_norm\_inputs                 &       False &         NaN &       False &       False &       False &          False &       False &          False \\
	rpf\_init\_weight\_scale           &        10.0 &        10.0 &        10.0 &        10.0 &        10.0 &           10.0 &        10.0 &           10.0 \\
	state\_expl\_idx                  &        None &        None &        [17] &      [0, 1] &        None &           None &        None &           None \\
	\bottomrule
\end{tabular}
}

\end{document}